\documentclass[nohyperref]{article}

\usepackage{microtype}
\usepackage{graphicx}
\usepackage{subfigure}
\usepackage{booktabs}

\usepackage[preprint]{neurips_2026}

\usepackage{hyperref}
 \hypersetup{
     colorlinks=true,
     linkcolor=citationcolor,
     filecolor=blue,
     citecolor=citationcolor,      
     urlcolor=cyan,
     }
\usepackage{url}            
\usepackage{booktabs}       
\usepackage{amsfonts}       
\usepackage{graphicx}
\usepackage{subfigure}
\usepackage{booktabs} 
\usepackage{amssymb}
\usepackage{amsmath}
\usepackage{amsthm}
\usepackage{mathtools}
\usepackage{tikz}
\usepackage{centernot}
\usepackage{caption}
\usepackage{lipsum}  
\usepackage{algorithm}
\usepackage{algorithmic}
\usepackage{etoolbox}\AtBeginEnvironment{algorithmic}{\small} 
\usepackage{wrapfig, blindtext}
\usepackage{multirow}
\usepackage{threeparttable}  
\usepackage{enumitem}

\newcommand{\ours}{\textbf{CoFi}}

\usepackage{cleveref}
\usepackage{xcolor,soul}
\usepackage{colortbl}

\newcommand{\calN}{{\mathcal{N}}}

\newcommand{\bbR}{\mathbb{R}}

\def\[#1\]{\begin{align}#1\end{align}}

\newcommand{\ie}{\textit{i}.\textit{e}., }
\newcommand{\eg}{\textit{e}.\textit{g}., }

\theoremstyle{plain}

\def\[#1\]{\begin{align}#1\end{align}}

\definecolor{myellow}{RGB}{194, 125, 47}
\definecolor{mgreen}{RGB}{48, 160, 111}

\definecolor{mteal}{RGB}{221, 254, 242}
\definecolor{bgteal}{RGB}{236, 245, 245}
\definecolor{mpurple}{RGB}{120, 111, 177}

\newcommand{\green}[1]{{\color{mgreen}{#1}}}

\newcommand{\purple}[1]{{\color{mpurple}{#1}}}

\usepackage{thmtools}
\usepackage{thm-restate}

\definecolor{citationcolor}{RGB}{80, 90, 180}

\newcommand{\ccolor}[1]{{\color{citationcolor}{#1}}}

\definecolor{mygray}{gray}{0.95}
\newcommand{\graybox}[1]{%
\begingroup
\setlength{\fboxsep}{0pt}%
\colorbox{mygray} {
\begin{minipage}{\linewidth}
\vspace{-0.5em}%
{#1}%
\end{minipage}%
}
\endgroup
}

\definecolor{mmgreen}{RGB}{243, 247, 243}
\definecolor{mmpurple}{RGB}{246, 242, 247}
\definecolor{mmccolor}{RGB}{235, 242, 250}

\definecolor{citationcolor}{RGB}{80, 90, 180}
\definecolor{background}{RGB}{240, 240, 250}
\definecolor{bggreen}{RGB}{229,242,229}
\newcommand{\cellbg}{\cellcolor{background}}
\newcommand{\cellte}{\cellcolor{mteal}}

\newcommand{\hlc}[2][yellow]{{%
    \colorlet{foo}{#1}%
    \sethlcolor{foo}\hl{#2}}%
}

\usepackage{amsmath,amsfonts,bm}

\def\eqref#1{(\ref{#1})}

\def\1{\bm{1}}

\DeclareMathAlphabet{\mathsfit}{\encodingdefault}{\sfdefault}{m}{sl}
\SetMathAlphabet{\mathsfit}{bold}{\encodingdefault}{\sfdefault}{bx}{n}

\title{Coarse-to-Fine Compositional Diffusion for Long-Horizon Planning}

\author{
    Byoungwoo~Park$^{1,2}$ \quad Utkarsh A. Mishra$^{2}$\quad Jaemoo Choi$^{2}$\quad Juho Lee$^{1}$\quad Yongxin Chen$^{2}$ \\
    $^{1}$KAIST \quad $^{2}$Georgia Institute of Technology \\\\
\centerline{\url{https://CoFi-diffusion.github.io/}}
}
\begin{document}

\maketitle

\begin{abstract}
Diffusion models provide strong priors for generating structured data, but many tasks require outputs beyond the scale on which these models are typically trained. Compositional generation addresses this by composing overlapping local plans from a pretrained short-horizon prior into a long-horizon output. However, standard composition primarily enforces agreement between neighboring local plans, yielding local consistency without directly specifying the global structure of the full composition. As a result, locally compatible plans may still form an implausible route, task sequence, or temporal evolution. Existing methods improve global coherence by repeatedly propagating local consistency signals or by adding inference-time optimization, but these procedures become expensive as the number or dimensionality of local plans increases. We propose Coarse-to-Fine Compositional Diffusion (CoFi), an inference-time sampler that separates global structure formation from local detail refinement. CoFi first aligns local denoised estimates around a shared coarse structure, producing a global scaffold that captures the long-range task-level arrangement. It then diffuses this scaffold to an intermediate noise level and denoises it with the same pretrained local prior, restoring local fine structure while preserving the scaffold-induced global coherence. Across long-horizon robotic planning, panoramic image generation, and long video generation, CoFi not only improves both global coherence and local sample quality over prior compositional baselines, but also requires $2$--$8\times$ fewer denoiser evaluations. 
\end{abstract}

\section{Introduction}

\begin{wrapfigure}[10]{r}{0.3\textwidth}
\vspace{-6mm}
\centering
\includegraphics[width=0.3\textwidth,]{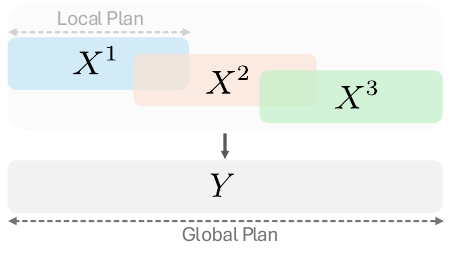}
\vspace{-4mm}
\caption{The global plan $Y$ is obtained by composing overlapping local plans $X^i$.}
\label{fig:figure_composite_concept} 
\end{wrapfigure}

Diffusion models provide strong priors for generating structured data~\citep{ho2020denoising, song2021denoising}, but many tasks require outputs beyond the scale on which these models are typically trained. For example, a robot may need to reach a distant goal by chaining short feasible motions into a long plan~\citep{luo2025generative}. Directly training a model for such long-horizon outputs is often impractical, since long-horizon data are scarce and the modeling cost grows quickly with sequence length or spatial extent~\citep{du2024position}. Compositional generation offers a practical alternative. Instead of training a long-horizon model, it generates overlapping local plans from a pretrained short-horizon prior and composes them into a single long-horizon plan~\citep{zhang2023diffcollage, mishra2023generative} as shown in~\Cref{fig:figure_composite_concept}. The same pretrained local prior can therefore be reused beyond its original generation scale, and the local plans can be generated in parallel.

\begin{figure}[!t]
\centering
\includegraphics[width=1.\textwidth,]{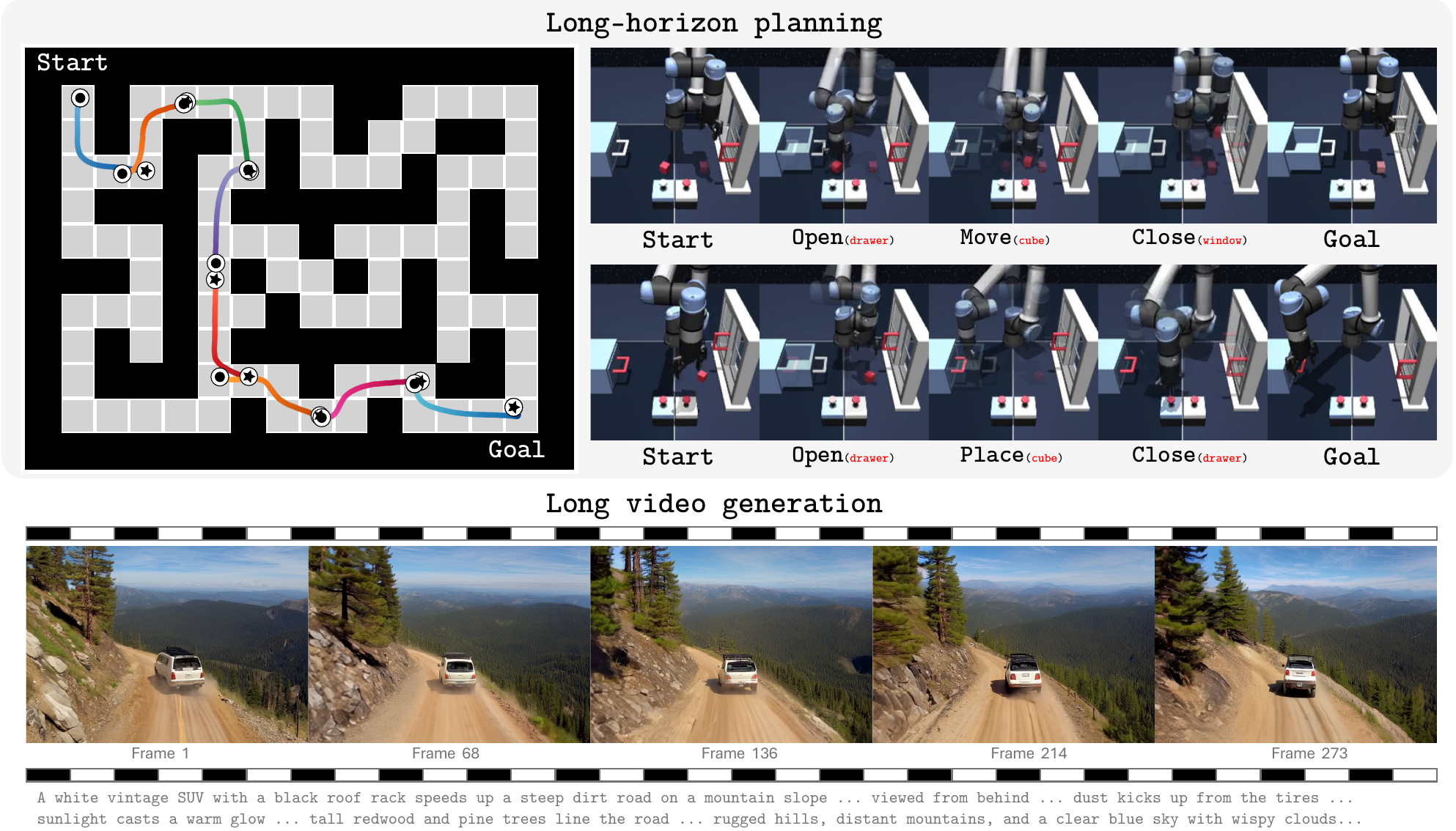}
\vspace{-4mm}
\caption{
\textbf{Scaling short-horizon priors by composition.}
A long structured global plan is built by composing overlapping local plans from a pretrained short-horizon prior. The goal is not only to enforce \green{\textit{local consistency}} (\eg smooth transitions between adjacent local plans), but also to preserve \purple{\textit{global coherence}} (\eg a plausible route, task sequence, or subject identity over the full horizon). We aim for a simple and effective inference-time sampler that satisfies both desired properties.
}
\label{fig:overall}
\vspace{-8mm}
\end{figure}

A typical approach to composition is to encourage neighboring local plans to agree at their shared overlaps during the reverse diffusion process~\citep{zhang2026compositional}. This local coupling reduces boundary mismatch and produces smooth transitions between adjacent local plans, which we refer to as \textit{local consistency}. However, \textit{local consistency} does not guarantee \textit{global coherence}: A composition can match every overlap and still lack a plausible task-level structure across the full horizon. For example, adjacent local plans may connect smoothly at their boundary yet follow incompatible routes that never reach the goal. In long videos, consecutive clips may align at their overlap while the subject’s identity gradually drifts over time. A compositional sampler must therefore satisfy a stronger requirement than boundary matching alone: adjacent local plans should remain mutually compatible, while their composition should induce a globally plausible long-horizon structure.

 Existing methods attempt to improve the global coherence by making the local consistency signal propagate more broadly across the local plans. Some methods repeat the overlap coupling multiple times within each denoising step, so that information can be passed from one segment to the next over longer distances~\citep{mishra2026compositional}. Other methods introduce inference-time optimization, often using a task-level objective or reward function, to directly adjust the local plans toward a more globally coherent composition~\citep{lee2023syncdiffusion,kim2025tuning}. While both strategies can improve \textit{global coherence}, they become expensive as the number of the local plans or the dimensionality of each segment increases. This limitation motivates the following design question:

\graybox{%
\

  \begin{center}
  \textit{Can we design a simple and effective compositional diffusion sampler that ensures both local consistency and global coherence without expensive inner iterations or optimization?}
  \end{center}
}

In this paper, we propose a simpler solution: instead of repeatedly propagating local consistency signals through neighboring overlaps, we first align the local denoised estimates around a \textit{shared coarse structure}. This stage exposes all local plans to a common global signal, promoting \textit{global coherence}, but it can also suppress plan-specific fine details. We therefore treat the resulting coarse structure as a scaffold rather than as the final sample. To restore the suppressed local structure, we diffuse the scaffold forward to an intermediate noise level and denoise it again with the same pretrained local prior, analogous to~\citep{meng2022sdedit}. Intuitively, the first stage determines the task-level arrangement of the composition, while the second stage restores local fine structure.

We call this procedure \textit{Coarse-to-Fine Compositional Diffusion} (\ours{}), an inference-time composition sampler for long-horizon generation with pretrained short-horizon priors. This two-stage design directly constructs a global scaffold, while adding only a partial second denoising pass for refinement. We evaluate CoFi on long-horizon robotic planning, panoramic image generation, and long video generation. Across these domains, CoFi improves both global coherence and local sample quality over prior compositional baselines, \textit{while requiring $2$--$8\times$ fewer denoiser evaluations} than baseline.

\section{Preliminaries}\label{sec:prelim}

\paragraph{Compositional generation} We consider the problem of constructing a long-horizon structured \textit{global plan} $Y \in \bbR^d$ from a collection of short-horizon \textit{local plans} $X \in \bbR^m$. Direct samples from the full long-horizon distribution $p(Y)$ are scarce, whereas short local behaviors are available and can be modeled reliably. We therefore assume access to a pretrained generative prior over local plans:
\[
 X^i = [x^{2i-1}, x^{2i}, x^{2i+1}], \quad i = 1, \dots, N,
\]
where $N$ is the number of plans. Here, the long-horizon structure arises through shared boundary factors between
consecutive plans. Let $\texttt{L}$ and $\texttt{R}$ denote linear selectors that extract
the \textit{left} and \textit{right} boundary factors of $X^i$, respectively:
\[
\texttt{L}(X^i) = x^{2i-1},\quad \texttt{R}(X^i) = x^{2i+1}, \quad \forall i = 1, \dots, N-1
\]
A valid composition requires neighboring local plans to agree on their shared \textit{overlapping} boundary,
\[
    \texttt{L} (X^{i+1}) = \texttt{R} (X^i)  =  x^{2i+1}, \quad \forall i = 1, \dots, N-1
\]
This induces a chain over sequence of \textit{the unique factors}, resulting the long-horizon global plan:
\[
Y = \texttt{Compose}(X^1, \dots, X^N) = [x^1, x^2, \dots, x^{2N+1}] .
\]
More generally, $\texttt{Compose}(\cdot) : \bbR^{m \times N} \to \bbR^d$ merges the overlapping local plans into a single global objective, either by using agreed boundary factors~\citep{luo2025generative} or by averaging overlapped regions~\citep{zhang2023diffcollage,mishra2023generative}. This merge step handles \textit{local consistency}, but it does not determine the global structure of the composed plan. Even when every adjacent pair agrees at the overlap, the full plan $Y$ can still drift over long horizons in style, identity, or task-level structure. We refer to this nonlocal requirement as \textit{global coherence}.

\paragraph{Local diffusion prior} In this paper, we focus on a diffusion prior for the local plan. Let $X_0 \in \mathbb{R}^m$ denote a local plan drawn from $p_\texttt{data}$. Following~\citep{ho2020denoising}, the forward process is given by
\[
q(X_t \mid X_0)
=
\mathcal{N}\!\left(
X_t;
\sqrt{\bar{\alpha}_t} X_0,
(1-\bar{\alpha}_t) I
\right),
\]
where $\bar{\alpha}_t := \prod_{s=1}^t (1-\beta_s)$, with $\beta_s \in (0,1)$. In other words, $X_t
=
\sqrt{\bar{\alpha}_t} X_0
+
\sqrt{1-\bar{\alpha}_t} \epsilon$ so that $X_t$ can be viewed as progressively interpolating between the clean local plan $X_0$ and Gaussian noise $\epsilon \sim \mathcal{N}(0,I)$. This places diffusion models within the broader view of stochastic interpolants and flow-based generative dynamics \citep{albergo2025stochastic, lipman2023flow}. The reverse process is approximated with a neural network $\epsilon_\theta(X_t,t) \approx - \sqrt{1-\bar{\alpha}_t} \nabla \log p_t(X_t)$ and sampling proceeds through a reverse transition parameterized by DDIM framework~\citep{song2021denoising}:
\[
& X_{t-1}
= \texttt{DDIM}_{\theta}(\hat{X}_{0|t}, t) :=
\sqrt{\bar{\alpha}_{t-1}} \hat{X}_{0 \mid t}
+
\sqrt{1-\bar{\alpha}_{t-1}-\sigma_t^2} \epsilon_\theta(X_t,t)
+
\sigma_t z,
\;\;
z \sim \mathcal{N}(0,I_{\scriptsize{m}}), \label{eq:DDIM} 
\\
& \hat{X}_{0 \mid t} :=
\tfrac{1}{\sqrt{\bar{\alpha}_t}}\left(X_t + (1-\bar{\alpha}_t) \nabla \log p_t(X_t) \right)
\approx
\tfrac{1}{\sqrt{\bar{\alpha}_t}}\left(X_t - \sqrt{1-\bar{\alpha}_t} \epsilon_\theta(X_t,t)\right) \label{eq:tweedie estimate}
\]
where $ \sigma_t
=
\eta
\sqrt{
\tfrac{1-\bar{\alpha}_{t-1}}{1-\bar{\alpha}_t}
}
\sqrt{
1-\tfrac{\bar{\alpha}_t}{\bar{\alpha}_{t-1}}
}.
$ This formulation is useful for compositional generation, since the reverse process admits direct inference-time modification through additional guidance terms.

It is worth noting that, if we apply the reverse transition in~\eqref{eq:DDIM} independently to each local plan, the induced model factorizes as $\prod_{i=1}^N p_t(X_t^i)$,
and ignores that the local plans should be different views of the same global plan $Y_t$. The resulting samples can be plausible as individual local plans, but still disagree at their overlaps or fail to form a coherent long-horizon object after composition.

\paragraph{Guided Compositional Diffusion} The ideal reverse process would sample from a locally consistent and yet globally coherent distribution over global plan $Y_t$, rather than from independent local priors $\prod_{i=1}^N p_t(X_t^i)$. In principle, one would like to guide the reverse process with an objective that captures both desired properties. A natural way to couple the local reverse processes is to tilt by a reward:
\[\label{eq:tilted_distribution}
p(Y_t)
\propto
e^{r_t(Y_t)} \textstyle{\prod_{i=1}^N} p(X_t^i),
\]
where composition reward $r_t$ penalizes invalid compositions. Ideally, this reward would capture both \textit{local consistency} among neighboring local plans $X_t^i$ and \textit{global coherence} of the composed global plan $Y_t$. However, designing such a reward is difficult because these requirements are qualitatively different: \textit{local consistency} is pairwise and overlap-based, whereas \textit{global coherence} is nonlocal and depends on the structure of the entire composition ($\ie$ plausible task-level plan).

\begin{wrapfigure}[16]{r}{0.35\textwidth}
    \centering
\includegraphics[width=0.35\textwidth,]{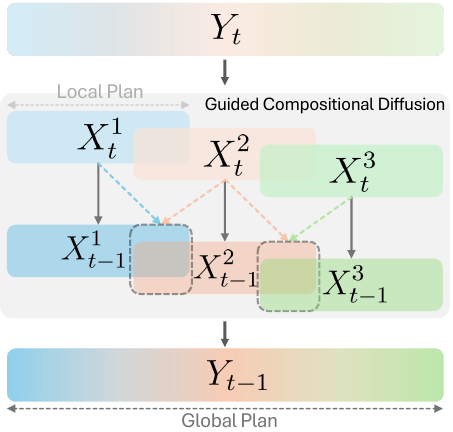}
\vspace{-6mm}
\caption{Local guidance reduces boundary mismatch but leaving the global coherence under-specified.}
\label{fig:figure_composite} 
\end{wrapfigure}

This makes the direct design of a tractable (and differentiable) compositional reward $r_t$ for both desired properties nontrivial, especially in high-dimensional plan spaces. For this reason, existing guided compositional samplers usually replace $r_t$ with local surrogates $r_t(Y_t) := \sum_{i=1}^N \hat r_t\left(X_t^i,X_t^{i+1}\right)$. For example, a composition technique~\citep{zhang2026compositional} imposes boundary agreement on the predicted clean plans $\hat{X}_{0|t}$ in~\eqref{eq:tweedie estimate}. In our notation, this corresponds to an overlap reward:
\[
    \hat r_t\left(X_t^i,X_t^{i+1}\right)
    =
    -\left\|
        \texttt{R}(\hat X^i_{0\mid t})
        -
        \texttt{L}(\hat X^{i+1}_{0\mid t})
    \right\|^2 .
\]
Additionally, the Bethe-style composition~\citep{zhang2023diffcollage, mishra2023generative} can be viewed as another local surrogate, where the gradient of the surrogate reward $\hat r_t$ is approximated using the two adjacent score:
\[\label{eq:bethe approximation}
\nabla \hat r_t\left(X_t^i,X_t^{i+1}\right) \approx
-\tfrac{1}{2}
\left[
\nabla \log p_t \left( \texttt{R}(X^i_t) | X_t^i \right)
+
\nabla \log p_t \left(\texttt{L}(X^{i+1}_t) |  X_t^{i+1} \right)
\right],
\]
These local surrogates provide a tractable guidance signal for the gradient of the global composition reward $r_t(Y_t)$ in~\eqref{eq:tilted_distribution}. In practice, the local reverse transition
in~\eqref{eq:DDIM} is guided by the surrogate reward:
\[\label{eq:guided_DDIM}
X^i_{t-1}
=
\texttt{DDIM}_{\theta}(\hat{X}^i_{0|t}, t)
+ \sigma_t \nabla \hat{r}_t(X_t^i, X_t^{i+1}),
\]
where $\texttt{DDIM}(\cdot, \cdot)$ denotes the unguided reverse transition in~\eqref{eq:DDIM}. This guidance is cheap and explicit, but it only communicates information through adjacent overlaps. It can reduce boundary mismatch, but it does not directly determine the coarse global structure of $Y_t$.

\section{Coarse-to-Fine Compositional Diffusion}\label{sec:method}

In this section, we introduce CoFi, \textit{a simple and effective inference-time diffusion framework for compositional generation} with pretrained local priors. CoFi follows a coarse-to-fine design: it first constructs \textit{a coarse global scaffold} that captures the long-range arrangement, and then refines this scaffold with the same pretrained local diffusion prior to recover \textit{fine local details}. 

Most tractable guidance objectives~\citep{mishra2023generative, zhang2026compositional} introduced in~\Cref{sec:prelim} act through local consistency as illustrated in~\Cref{fig:figure_composite}, with canonical choices such as the Bethe approximation in~\eqref{eq:bethe approximation}. They encourage neighboring local plans to agree, but they do not directly specify whether the full composition has a plausible task-level structure. A single local guidance step in~\eqref{eq:guided_DDIM} only couples adjacent overlaps, leaving distant local plans weakly coordinated. Recent methods address this limitation by repeating the local coupling inside the sampling process. For example, CDGS~\citep{mishra2026compositional} improves global coherence through inner iterations. These steps can be viewed as a form of iterative message passing: information first moves between neighboring local plans, and repeated updates gradually expand the effective communication range across the chain. This is effective, but the cost grows with the number of local plans and becomes especially large for high-dimensional compositions such as visual planning.

In this paper, we aim to develop a simple procedure that remains computationally efficient while satisfying both local consistency and global coherence. To do so, we replace repeated local message passing with a shared scaffold over the local denoised estimates. This gives all local plans access to a common coarse signal, forming a global scaffold. Because the global arrangement has already been outlined, refinement only needs to restore local fine structure rather than propagate global constraints again. We therefore diffuse the scaffold forward to an intermediate noise level $t^{\star} \in \{1,2,\ldots,T\}$ and denoise it again using the local guided step in~\eqref{eq:guided_DDIM}. This diffusion refinement step follows the same principle as~\citep{meng2022sdedit}: keep the coarse structure, then let the diffusion prior restore details. 

\paragraph{Coarse global plan construction.} 
We first define the scaffold alignment update that produces the coarse global scaffold. For each denoising step $t$, the clean estimate for each local plan $\green{\hat X^i_{0| t}}$ may be individually plausible, yet still fail to induce a coherent global composition. We therefore seek a coarse estimate that retains the shared global structure while attenuating plan-specific variation. 

To do so, the coarse estimate should satisfy two competing requirements: for each local plan $X^i$, it should remain close to the denoised estimate $\green{\hat X^i_{0| t}}$ so that local plausibility is preserved, while at the same time admitting a decomposition into a shared global component and plan-specific residuals. To formalize this trade-off, we define a simplified scaffold alignment objective for each local plan:
\graybox{
\[\label{eq:consensus minimization}
\min_{Z^1, \cdots, Z_N}
\sum_{i=1}^N \left[ 
\|Z^i-\green{\hat X^i_{0| t}}\|^2
+ \gamma_t \|Z^i-\bar Z\|^2
\right],
\qquad
\bar Z := \frac{1}{N} \textstyle{\sum_{i=1}^N} Z^i
\]
}

where $\gamma_t$ controls the trade-off between local fidelity and shared component. The first term preserves local plausibility, whereas the second promotes a shared global scaffold across local plans. Since \eqref{eq:consensus minimization} is a quadratic problem, its minimizer admits the closed-form solution below, which interpolates each local estimate toward the shared scaffold. Let $Z^{i,\star} := \purple{\bar X^i_{0| t}}$ be the minimizer of \eqref{eq:consensus minimization}. Then:
\[\label{eq:shared consensus}
\purple{\bar X^i_{0| t}}
=
(1 - \lambda_t) \bar X_{0|t}
+
\lambda_t\green{\hat X^i_{0|t}}, \quad \bar X_{0|t}:= \frac{1}{N}  \textstyle{\sum_{i=1}^N}  \green{\hat X^i_{0|t}}
\]
where $\lambda_t := \tfrac{1}{1+\gamma_t} \in [0, 1]$. If $\lambda_t=1$ (\ie $\gamma_t = 0$ in~\eqref{eq:consensus minimization}), the scaffold term has no effect and $\purple{\bar X^i_{0| t}}=\green{\hat X^i_{0| t}}$. If $\lambda_t=0$ (\ie $\gamma_t \rightarrow \infty$ in~\eqref{eq:consensus minimization}), all local estimates collapse to the shared scaffold $\bar X_{0| t}$. In other words, each optimal coarse estimate $\purple{\bar X^i_{0| t}}$ is obtained by shrinking the local denoised estimate $\green{\hat X^i_{0| t}}$ toward the shared scaffold mean. The shared global component $\bar X_{0| t}$ is preserved, while the plan-specific residual $\green{\hat X^i_{0| t}}-\bar X_{0| t}$ is attenuated by a factor of $\lambda_t$. We then replace the local denoised estimate $\green{\hat X^i_{0| t}}$ in \eqref{eq:guided_DDIM} with scaffold alignment $\purple{\bar X^i_{0| t}}$ for global coherency:
\[\label{eq:coarse reverse process}
X^i_{t-1}
= \underbrace{\texttt{DDIM}_{\theta}(\purple{\bar X^i_{0| t}}, t)}_{\texttt{global}}
+ \underbrace{\sigma_t \nabla \hat{r}_t(X_t^i, X_t^{i+1})}_{\texttt{local}},
\]
As a result, the reverse process in~\eqref{eq:coarse reverse process} is biased toward a globally aligned but locally coarse scaffold. This is precisely the role of the coarse global plan: it suppresses incompatible local variation while preserving the long-range structure shared across local plans, leaving fine local detail to the subsequent refinement stage. Repeating this update from
$t=T$ to $t=0$ gives a coarse-structured global plan:
\[\label{eq:composed global plan}
Y^{\texttt{c}}_0
=
\texttt{Compose}(X^{1}_0,\ldots,X^{N}_0).
\]
Here, the superscript $\mathrm{c}$ denotes the coarse stage. The resulting coarse global plan $Y^{\texttt{c}}_0$ captures the global arrangement of the composition, but may lack fine local detail because the scaffold construction stage softens local fine details by the timestep-dependent factor $\lambda_t$.

\begin{figure}[!t]
\begin{minipage}[t]{0.47\textwidth}
  \begin{algorithm}[H]
    \caption{Coarse Scaffold Construction}
    \label{alg:coarse_plan}
    \begin{algorithmic}
    \STATE \textbf{Input:} local denoiser $\epsilon_\theta$, local guidance reward $\hat r_t$

    \hspace*{-\fboxsep}\colorbox{background}{\parbox{\linewidth}{%
    \STATE Initialize local plans $X_T^1,\ldots,X_T^N \sim \mathcal N(0,I)$
      \FOR{$t=T, \cdots, 1$}
        \STATE \textbf{for} $i=1, \cdots, N$ \textbf{do in parallel}
          \STATE\hspace{+4mm} Compute Tweedie estimate $\hat X^i_{0| t}$ using~\eqref{eq:tweedie estimate}
          \STATE\hspace{+4mm} Estimate $\bar{X}^i_{0| t}$ using~\eqref{eq:shared consensus}
          \STATE\hspace{+4mm} Update $X^i_{t-1}$ using guided step in~\eqref{eq:coarse reverse process}
        \STATE \textbf{end for}
      \ENDFOR
    \STATE Compose coarse global plan $Y^{\texttt{c}}_0$ using~\eqref{eq:composed global plan}.
    }}
    \STATE \textbf{Output:} coarse global plan $Y^{\texttt{c}}_0$
    \end{algorithmic}
  \end{algorithm}
\end{minipage}
\hfill
\begin{minipage}[t]{0.51\textwidth}
  \begin{algorithm}[H]
    \caption{Structure-Preserving Refinement}
    \label{alg:diffusion_refinement}
    \begin{algorithmic}
    \STATE \textbf{Input:} coarse global plan $Y^{\texttt{c}}_0$, refinement timestep $t^\star$, local denoiser $\epsilon_\theta$, local guidance reward $\hat r_t$

    \hspace*{-\fboxsep}\colorbox{bgteal}{\parbox{\linewidth}{%
    \STATE Get noised scaffold $Y^{\texttt{r}}_{t^\star}$ by noising $Y^{\texttt{c}}_{0}$ using~\eqref{eq:structure_preserving_noise}
    \STATE Let $\{X^{i,\texttt{r}}_{t^\star}\}_{i=1}^N$ be the local views induced by $Y^{\texttt{r}}_{t^\star}$
      \FOR{$t=t^\star,\cdots,1$}
        \STATE \textbf{for} $i=1, \cdots, N$ \textbf{do in parallel}
          \STATE\hspace{+4mm} Compute Tweedie estimate $\hat X^{i,\texttt{r}}_{0| t}$ using~\eqref{eq:tweedie estimate}
          \STATE\hspace{+4mm} Update $X^{i,\texttt{r}}_{t-1}$ using the guided step in~\eqref{eq:guided_DDIM}
        \STATE \textbf{end for}
      \ENDFOR
    \STATE Compose refined global plan $Y^{\texttt{r}}_0$ using~\eqref{eq:composed global plan}.
        }}
        
    \STATE \textbf{Output:} final global plan $Y^{\texttt{r}}_0$
    \end{algorithmic}
  \end{algorithm}
\end{minipage}
\vspace{-4mm}
\end{figure}

\paragraph{Structure-preserving diffusion refinement.}

Starting from the coarse global plan $Y^{\texttt{c}}_0$, we re-inject noise at an intermediate timestep $t^{\star}$ and run a second reverse diffusion pass. The purpose of this stage is to recover locally plausible detail under the pretrained diffusion prior while preserving the global arrangement induced by the coarse scaffold. More concretely, the coarse global plan determines \textit{where} the global structure should lie, whereas the second reverse diffusion pass determines \textit{how} the final composition should appear at the local level. Formally, after the coarse stage produces the global plan $Y^{\texttt{c}}_0$ in~\eqref{eq:composed global plan}, we diffuse it forward to an intermediate timestep $t^\star \in \{1,2,\ldots,T\}$:
\[\label{eq:structure_preserving_noise}
Y^{\texttt{r}}_{t^\star}
=
\sqrt{\bar\alpha_{t^\star}}Y^{\texttt{c}}_0
+
\sqrt{1-\bar\alpha_{t^\star}}\varepsilon,
\quad
\varepsilon \sim \mathcal N(0,I_{\scriptsize{d}}).
\]
The forward noising step in~\eqref{eq:structure_preserving_noise} is applied to the coarse global scaffold $Y^{\texttt{c}}_0$, rather than separately to each local plan $X_0^i$. If the local plans were noised independently, each plan would receive its own perturbation, which could break the alignment produced by the scaffold construction stage in~\eqref{eq:coarse reverse process}. Noising the global scaffold with a shared noise $\varepsilon$ in~\eqref{eq:structure_preserving_noise} avoids this issue: the local plans are induced from the same prior $Y^{\texttt{r}}_{t^\star}$ and start as compatible views of a shared global plan. It restores local fine structure~\citep{meng2022sdedit} while keeping the reverse process anchored to the global plan in~\eqref{eq:composed global plan}.

After this structure-preserving noising step, we continue the reverse process locally. Let $X^{i,\texttt{r}}_{t^{\star}}$ denote the $i$-th local plan induced by the refined global plan $Y^{\texttt{r}}_{t^{\star}}$. Starting from $t^\star$, each local plan follows the guided composition in~\eqref{eq:guided_DDIM}. We summarize the two-stage sampling procedure in Algorithms~\ref{alg:coarse_plan} $\&$~\ref{alg:diffusion_refinement}.
\section{Experiments}

We evaluate CoFi on three compositional generation domains: (1)~long-horizon robotic planning, (2)~panoramic image generation, and (3)~long video generation. We adopt the experimental protocol of \citet{mishra2026compositional} and use the same pretrained diffusion models, datasets, and evaluation metrics.

\begin{itemize}[leftmargin=10pt]
\item \textit{Robotic planning.}\;\; 
We evaluate on OGBench~\citep{park2025ogbench}, which includes stitch and play datasets for \texttt{PointMaze}, \texttt{AntMaze}, and \texttt{Scene} manipulation tasks. The stitch datasets contain short maze trajectories, while the play datasets contain unstructured interaction data. Neither directly solves the evaluation tasks, so successful planning requires composing short fragments into a longer plan. We report success rates over $100$ episodes averaged across $5$ seeds. For maze tasks, we follow \citet{luo2025generative} and use distance-based replanning.

\item \textit{Panoramic images.}\;\;
We use Stable Diffusion~2.0~\citep{rombach2022high} as the local image prior. Each panorama is generated by composing $9$ overlapping $512\!\times\!512$ patches into a $512\!\times\!4608$ image with $50\%$ overlap. We evaluate $14$ prompts and report Intra-LPIPS~\citep{zhang2018unreasonable}, Intra-Style loss~\citep{gatys2016image}, and CLIP score~\citep{radford2021learning}.

\item \textit{Long videos.}\;\;
We use CogVideoX-2B~\citep{yang2025cogvideox} as the local video prior. The base model natively generates short clips of $50$ frames, and we extend it to $273$ frames at $720$p by composing $9$ temporal chunks with $50\%$ overlap. We evaluate $8$ prompts using VBench~\citep{huang2024vbench}, reporting subject consistency, temporal flickering, aesthetic quality, and prompt alignment.
\end{itemize}

\begin{table}[t]
\caption{\textbf{Quantitative results on compositional planning.}
We report OGBench~\citep{park2025ogbench} success rates over $100$ evaluation episodes averaged across $5$ seeds. \hlc[background]{Best results are highlighted}.}
\centering
\resizebox{\textwidth}{!}{%
\renewcommand{\arraystretch}{1.2}
\setlength{\tabcolsep}{6pt}
\begin{tabular}{@{} l l c c c c c c c c c @{}}
\toprule
Env 
& Size & GCBC & GCIVL & GCIQL & HIQL & GSC & CD & CDGS &  \ours{}  \\
\midrule
\multirow{3}{*}{\textbf{\texttt{\shortstack[l]{Pointmaze \\ \ccolor{Stitch}}}}} 
& \textbf{\texttt{Medium}} 
& 23{\color{gray}\scriptsize$\pm$18} 
& 70{\color{gray}\scriptsize$\pm$14} 
& 21{\color{gray}\scriptsize$\pm$9} 
& 74{\color{gray}\scriptsize$\pm$6}  
& \cellbg 100{\color{gray}\scriptsize$\pm$0} 
& \cellbg 100{\color{gray}\scriptsize$\pm$0}
& \cellbg 100{\color{gray}\scriptsize$\pm$0}
& \cellbg 100{\color{gray}\scriptsize$\pm$0} \\
& \textbf{\texttt{Large}} 
& 7{\color{gray}\scriptsize$\pm$5}   
& 12{\color{gray}\scriptsize$\pm$6} 
& 31{\color{gray}\scriptsize$\pm$2}  
& 13{\color{gray}\scriptsize$\pm$6} 
& \cellbg 100{\color{gray}\scriptsize$\pm$0} 
& \cellbg 100{\color{gray}\scriptsize$\pm$0}
& \cellbg 100{\color{gray}\scriptsize$\pm$0}
& \cellbg 100{\color{gray}\scriptsize$\pm$0} \\
& \textbf{\texttt{Giant}}  
& 0{\color{gray}\scriptsize$\pm$0}  
& 0{\color{gray}\scriptsize$\pm$0}  
& 0{\color{gray}\scriptsize$\pm$0} 
& 0{\color{gray}\scriptsize$\pm$0} 
& 29{\color{gray}\scriptsize$\pm$3}  
& 68{\color{gray}\scriptsize$\pm$3} 
& 82{\color{gray}\scriptsize$\pm$4}
& \cellbg 96{\color{gray}\scriptsize$\pm$2} \\
\midrule
\multirow{3}{*}{\textbf{\texttt{\shortstack[l]{Antmaze \\ \ccolor{Stitch}}}}} 
& \textbf{\texttt{Medium}} 
& 45{\color{gray}\scriptsize$\pm$11} 
& 44{\color{gray}\scriptsize$\pm$6} 
& 29{\color{gray}\scriptsize$\pm$6} 
& 94{\color{gray}\scriptsize$\pm$1}  
& 96{\color{gray}\scriptsize$\pm$2} 
& \cellbg  97{\color{gray}\scriptsize$\pm$2}
& \cellbg  97{\color{gray}\scriptsize$\pm$1}
& \cellbg 97{\color{gray}\scriptsize$\pm$2} \\
& \textbf{\texttt{Large}} 
& 3{\color{gray}\scriptsize$\pm$3}   
& 18{\color{gray}\scriptsize$\pm$2} 
& 7{\color{gray}\scriptsize$\pm$2}  
& 67{\color{gray}\scriptsize$\pm$5} 
& 66{\color{gray}\scriptsize$\pm$2} 
& 86{\color{gray}\scriptsize$\pm$2}
& \cellbg 88{\color{gray}\scriptsize$\pm$2}
& \cellbg 88{\color{gray}\scriptsize$\pm$2}  \\
& \textbf{\texttt{Giant}}  
& 0{\color{gray}\scriptsize$\pm$0}  
& 0{\color{gray}\scriptsize$\pm$0}  
& 0{\color{gray}\scriptsize$\pm$0} 
& 21{\color{gray}\scriptsize$\pm$2} 
& 20{\color{gray}\scriptsize$\pm$1}  
& 65{\color{gray}\scriptsize$\pm$3} 
& 84{\color{gray}\scriptsize$\pm$3}
& \cellbg 85{\color{gray}\scriptsize$\pm$3} \\
\midrule
\textbf{\texttt{Scene}} \textbf{\texttt{\ccolor{Play}}}
& -
& 5{\color{gray}\scriptsize$\pm$1}
& 42{\color{gray}\scriptsize$\pm$4}
& 51{\color{gray}\scriptsize$\pm$4}
& 38{\color{gray}\scriptsize$\pm$3}
& 8{\color{gray}\scriptsize$\pm$2}
& 13{\color{gray}\scriptsize$\pm$1}
& 51{\color{gray}\scriptsize$\pm$2}
& \cellbg 63{\color{gray}\scriptsize$\pm$3} \\
\bottomrule
\end{tabular}%
}
\label{tab:robotics}
\vspace{-3mm}
\end{table}

\paragraph{Baselines.}
We compare with both non-compositional and compositional baselines. For robotic planning, we include goal-conditioned methods, GCBC~\citep{lynch2020learning,ghosh2021learning}, GCIVL, GCIQL~\citep{kostrikov2021offline}, and HIQL~\citep{park2023hiql}, as well as compositional methods, GSC~\citep{mishra2023generative}, CD~\citep{luo2025generative}, and CDGS~\citep{mishra2026compositional}. For panoramic image generation, we compare with Multi-Diffusion~\citep{bar2023multidiffusion} (the image-domain analogue of GSC), SyncDiffusion~\citep{lee2023syncdiffusion}, and CDGS. For long video generation, we compare with Gen-L-Video~\citep{wang2023gen} (the video-domain analogue of GSC), and CDGS. We also report CogVideoX-2B at its native short horizon of $50$ frames as a video reference.

\paragraph{Compositional long-horizon robotic planning.}
\Cref{tab:robotics} reports success rates on OGBench~\citep{park2025ogbench}. The medium and large \texttt{PointMaze} settings are saturated by existing compositional methods, so the harder Giant and Scene-Play tasks are more informative. On \texttt{PointMaze-Giant}, we achieve $96\%$ success, improving over CDGS by $14\%$, suggesting that the scaffold construction stage helps resolve long-range inconsistency in maze-like trajectory composition. On \texttt{AntMaze-Giant}, we obtain $85\%$, comparable to CDGS at $84\%$. 
\begin{wrapfigure}[13]{r}{0.6\textwidth}
\vspace{-4mm}
    \centering
\begin{minipage}[b]{0.3\linewidth}
    \centering
    \scriptsize
    {\texttt{Scaffold}}
\end{minipage}
\begin{minipage}[b]{0.33\linewidth}
    \centering
    \scriptsize
    {\texttt{Renoise}}
\end{minipage}
\begin{minipage}[b]{0.3\linewidth}
    \centering
    \scriptsize
    {\texttt{Refinement}}
\end{minipage}
\includegraphics[width=0.6\textwidth,]{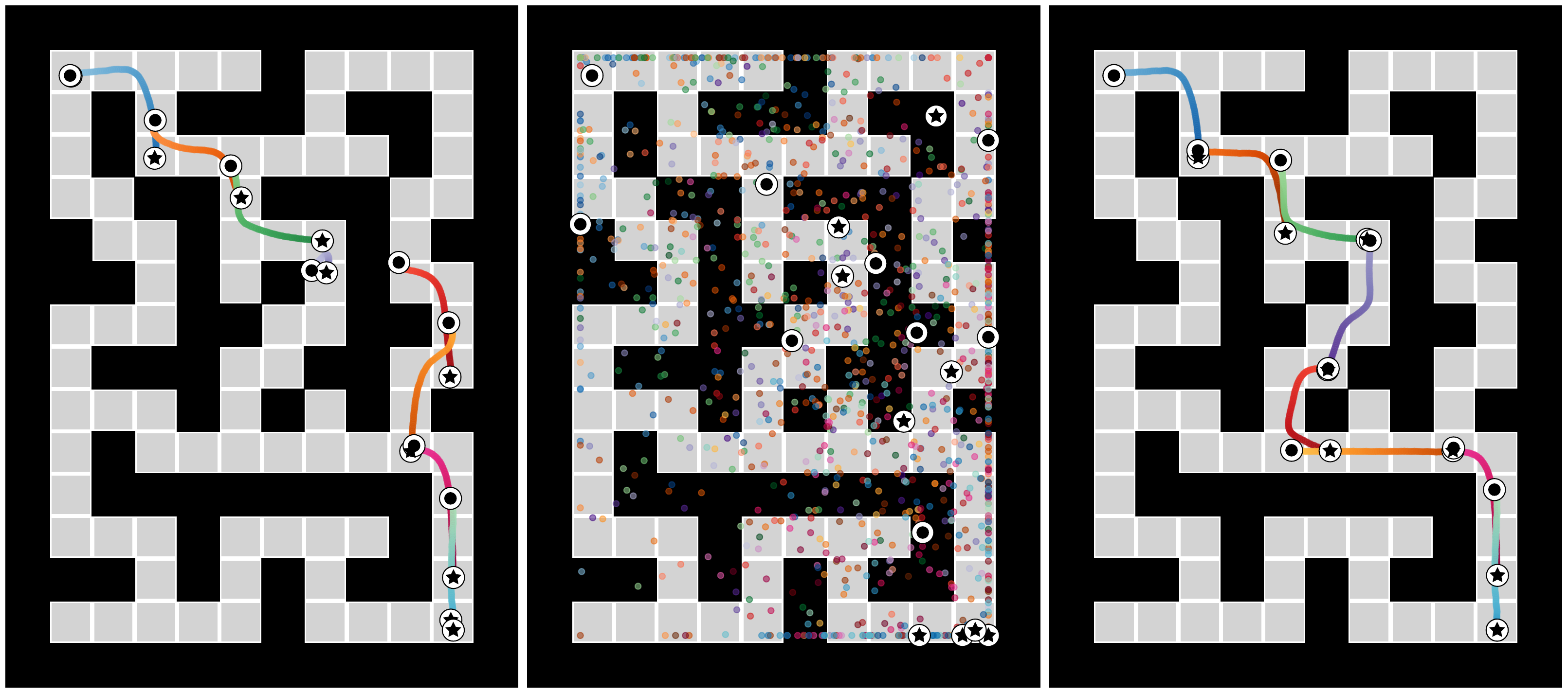}
\vspace{-5mm}
\caption{\textbf{Robotic Planning.}
A coarse scaffold route is refined into a plausible task-level trajectory.}
\label{fig:figure_maze} 
\end{wrapfigure}
Here, the strongest compositional baselines already find reasonable global routes, and the remaining failures likely come from inverse-dynamics tracking~\citep{luo2025generative}. On \texttt{Scene-Play}, we improve over CDGS from $51\%$ to $63\%$, where the gain suggests that the coarse scaffold stage provides a more stable scaffold for highly multimodal play data. Non-compositional baselines degrade sharply in the harder long-horizon settings. For example, on \texttt{PointMaze-Giant}, while compositional methods remain effective compared to non-compositional baselines obtain $0\%$ success. \Cref{fig:figure_maze} shows a generated plan from CoFi on \texttt{PointMaze-Giant}. The scaffold construction stage captures the global route but may still produce wall-crossing artifacts. The subsequent refinement stages refine the plan into a globally aligned trajectory.

\begin{figure}[!t]
\centering
\includegraphics[width=1.\textwidth,]{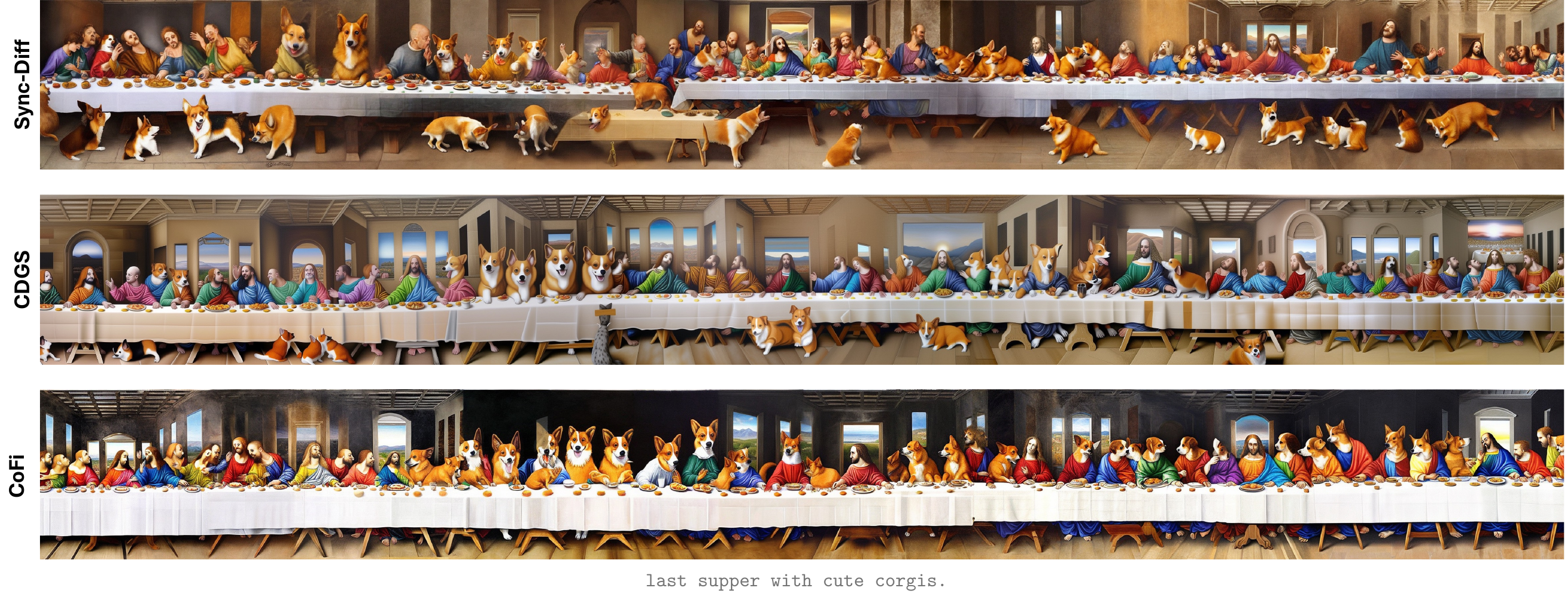}
\vspace{-6mm}
\caption{\textbf{Panoramic image generation.} The coarse-to-fine procedure maintains a consistent global style across distant chunks. Baselines show abrupt changes or gradual style drift over the panorama.}
\label{fig:panorama}
\vspace{-7mm}
\end{figure}

\begin{wraptable}[15]{r}{0.53\textwidth}
\vspace{-4mm}
\caption{\textbf{Quantitative results on panoramic image generation.}
We compose $9$ overlapping $512 \times 512$ local patches into a $512 \times 4608$ panorama and evaluate patch consistency and text alignment on generated panorama. \hlc[background]{Best results are highlighted}.}
\vspace{-2mm}
\centering
\renewcommand{\arraystretch}{1.2}
\setlength{\tabcolsep}{6pt}
\begin{tabular}{@{} l c c c c c c c c c @{}}
\toprule
& \multicolumn{2}{c}{{\color{gray} \texttt{\textbf{Style}}}} & {\color{gray} \texttt{\textbf{Semantic}}} \\
\cmidrule(l){2-3}  \cmidrule(l){4-4}
Method & LPIPS $\downarrow$ & Style-L $\downarrow$ & CLIP-S $\uparrow$ \\
\midrule
GSC {\color{gray}\tiny{(Multi-Diff)}}
& 0.72{\color{gray}\scriptsize$\pm$0.08}
& 2.96{\color{gray}\scriptsize$\pm$0.24}
& 31.77{\color{gray}\scriptsize$\pm$2.14}   \\
Sync-Diff
& 0.60{\color{gray}\scriptsize$\pm$0.05}
& 1.36{\color{gray}\scriptsize$\pm$1.07}
& 32.40{\color{gray}\scriptsize$\pm$1.85}   \\
CDGS
& 0.65{\color{gray}\scriptsize$\pm$0.05}
& 1.71{\color{gray}\scriptsize$\pm$1.03}
& 32.67{\color{gray}\scriptsize$\pm$1.92}   \\
\midrule
\ours $\;${\color{gray}\tiny{(Ours)}}
& \cellbg 0.48{\color{gray}\scriptsize$\pm$0.08}
& \cellbg 0.72{\color{gray}\scriptsize$\pm$0.54}
& \cellbg 33.18{\color{gray}\scriptsize$\pm$1.84}   \\
\bottomrule
\end{tabular}%
\label{tab:panorama}
\end{wraptable}

\paragraph{Panoramic image generation.}

\Cref{tab:panorama} and \Cref{fig:panorama} summarize the panorama generation results. We achieve the best scores
across all three metrics among the compared methods. In particular, relative to CDGS, our method reduces Intra-LPIPS from $0.65$ to $0.48$ and Intra-Style loss from $1.71$ to $0.72$, while also improving CLIP score from $32.67$ to $33.18$. These gains indicate that the generated patches remain more consistent in color, texture, and overall style across the full panorama, without sacrificing text-image alignment. Qualitatively, our method produces panoramas with a more stable global style and fewer abrupt transitions across patch boundaries. Unlike SyncDiffusion, which relies on an additional perceptual optimization loop during sampling, our method achieves strong coherence through the same coarse-to-fine procedure used in the planning experiments. This indicates that the proposed method improves global coherence while preserving prompt-relevant local details and semantic fidelity.

\begin{figure}[!t]
\centering
\includegraphics[width=1.\textwidth,]{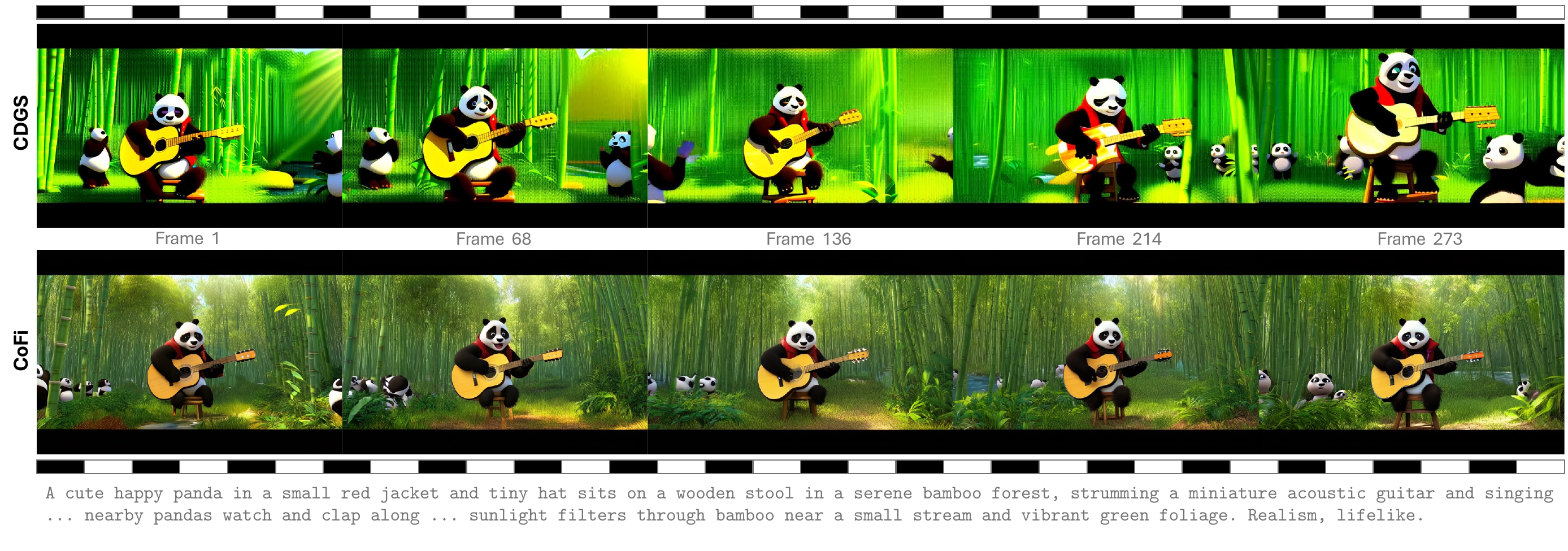}
\vspace{-6mm}
\caption{\textbf{Long video generation.} The coarse-to-fine procedure preserves subject appearance and scene structure over long temporal ranges. Baseline shows temporal drift along distant frames.}
\vspace{-4mm}
\label{fig:video}
\end{figure}

\paragraph{Long video generation.}
\Cref{tab:video} and \Cref{fig:video} report the long video generation results. We improve subject consistency from $91.67$ with CDGS to $94.11$, approaching the native $50$-frame CogVideoX reference at $95.91$. This is the key metric in this setting, since temporal composition often fails through subject drift over long horizons. We also improve aesthetic quality from $58.90$ with CDGS to $62.36$, close to the native short-horizon reference at $63.10$, and obtain the best prompt alignment among compositional baselines at $26.86$. These results suggest that the refinement stage recovers local visual detail after the coarse global structure is fixed, rather than relying on expensive search throughout sampling. Temporal flickering is similar across compositional methods. Although the native $273$-frame CogVideoX extension obtains the highest flickering score, this metric does not fully capture long-range subject drift or semantic inconsistency. As shown in~\Cref{fig:video}, the coarse-to-fine procedure better preserves subject appearance and scene layout across the full video.

\begin{table}[t]
\caption{\textbf{Quantitative results on long video generation.}
We compose $50$-frame short-horizon CogVideoX-2B videos into a $273$-frame long-horizon video and evaluate Vbench on generated long-horizon videos. The $50$-frame video serves as \hlc[mteal]{a reference}. \hlc[background]{Best results are highlighted}.}
\centering
\renewcommand{\arraystretch}{1.2}
\setlength{\tabcolsep}{6pt}
\begin{tabular}{@{} l c c c c c c c c c @{}}
\toprule
& \multicolumn{2}{c}{{\color{gray} \texttt{\textbf{Temporal}}}} & {\color{gray} \texttt{\textbf{Frame}}} & {\color{gray} \texttt{\textbf{Semantic}}} \\
\cmidrule(l){2-3}  \cmidrule(l){4-4}  \cmidrule(l){5-5}
Method & Subject $\uparrow$ & Flickering $\uparrow$ & Aesthetic $\uparrow$ & Prompt $\uparrow$ \\
\midrule
CogVideoX {\color{gray}\tiny{50frames}}
& \cellte 95.91
& \cellte 97.36
& \cellte 63.10
& \cellte 25.51 \\
CogVideoX {\color{gray}\scriptsize{273frames}}
& 90.24
& 98.44
& 49.44
& 21.78 \\
GSC {\color{gray}\tiny{(Gen-L-Video)}}
& 89.51
& 96.89
& 60.12
& 25.13 \\
CDGS
& 91.67
& 97.16
& 58.90
& 26.13 \\
\midrule
\ours $\;${\color{gray}\tiny{(Ours)}}
& \cellbg 94.11
& \cellbg 97.27
& \cellbg 62.36
& \cellbg 26.86 \\
\bottomrule
\end{tabular}%
\label{tab:video}
\vspace{-5mm}
\end{table}

\paragraph{Inference cost.}

\begin{wrapfigure}[13]{r}{0.6\textwidth}
\vspace{-5mm}
    \centering
\begin{minipage}[b]{0.3\linewidth}
    \centering
    \scriptsize
    {\texttt{Planning}}
\end{minipage}
\begin{minipage}[b]{0.33\linewidth}
    \centering
    \scriptsize
    {\texttt{Image}}
\end{minipage}
\begin{minipage}[b]{0.3\linewidth}
    \centering
    \scriptsize
    {\texttt{Video}}
\end{minipage}
\includegraphics[width=0.6\textwidth,]{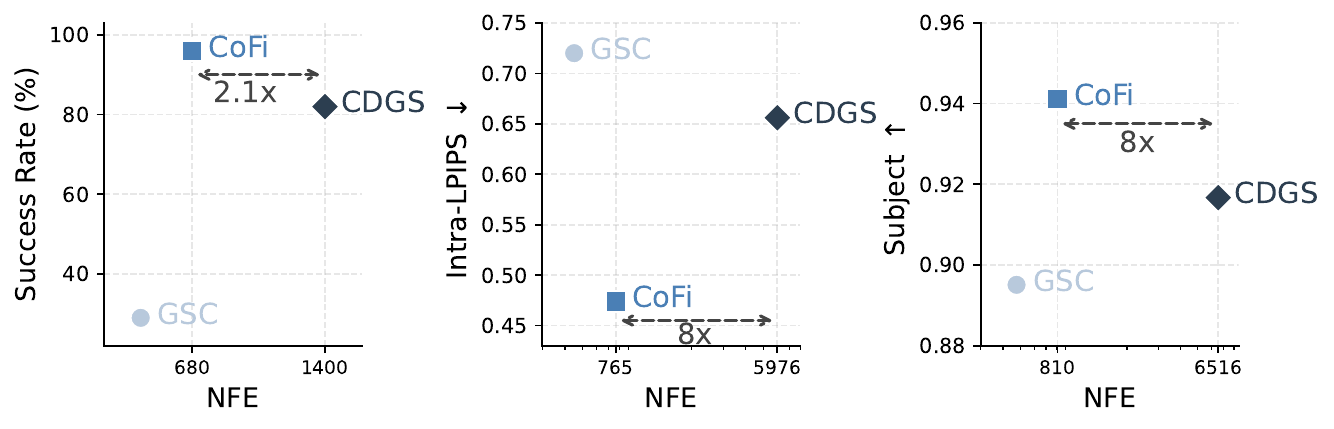}
\vspace{-6mm}
\caption{\textbf{NFE--performance Pareto comparison.}
We compare performance against the NFE across the three domains. CoFi improves CDGS with \textbf{$\mathbf{2}$--$\mathbf{8\times}$ fewer NFE}, substantially reducing computational overhead while improving global coherence and local sample quality.}
\label{fig:nfe_comparison} 
\end{wrapfigure}

\Cref{fig:nfe_comparison} compares the number of function evaluations (NFE) across the three domains. GSC is the cheapest because it uses local guidance without resampling or search, but the global structure remains under-specified. CDGS adds inner iterations, population search, and pruning to improve global coherence, which makes the cost grow quickly as the number of local plans increases. This is especially costly for panorama and video generation, where many high-dimensional chunks must be coordinated. In contrast, CoFi separates global and local refinement. The scaffold construction stage uses the same denoiser evaluations as GSC, and the refinement stage adds only $t^\star$ extra steps. The total cost is therefore $T+t^\star$, giving a small overhead over GSC while requiring $2$--$8\times$ fewer function evaluations than CDGS.

\paragraph{Ablation Study.}

\begin{figure}[t]
\begin{minipage}[t]{0.7\textwidth}
\includegraphics[width=1.\textwidth,]{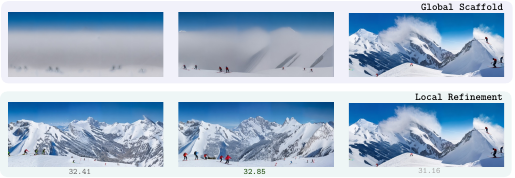}
\vspace{-6mm}
\caption{\textbf{Effect of scaffold schedule.}
We compare strong scaffold alignment (left) with constant $\lambda_t=0.8$, (middle) the scheduled $\lambda_t$ in Figure~\ref{fig:lambda_graph}, and (right) no scaffold alignment with constant $\lambda_t=1$.
The top row shows the coarse scaffold $Y_0^{\texttt{c}}$, and the bottom row shows the refinement $Y_0^{\texttt{r}}$. Numbers below the images report the CLIP score.}
\label{fig:lambda_ablation} 
\end{minipage}
\centering
\hfill
\begin{minipage}[t]{0.28\textwidth}
\vspace{-34mm}
\includegraphics[width=1.\textwidth,]{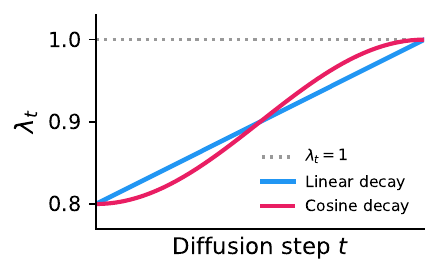}
\vspace{-6mm}
\caption{Smaller $\lambda_t$ gives stronger scaffold alignment. We use stronger alignment at early diffusion steps to align local plans around a shared coarse mode, and gradually increase $\lambda_t \rightarrow 1$.}
\label{fig:lambda_graph} 
\end{minipage}

\end{figure}

We study the role of the scaffold alignment coefficient $\lambda_t$ in~\eqref{eq:coarse reverse process}. Smaller $\lambda_t$ pulls local denoised estimates more strongly toward the shared scaffold, whereas $\lambda_t=1$ leaves them unchanged. As shown in~\Cref{fig:lambda_ablation}, a fixed $\lambda_t=0.8$ over-smooths the scaffold, while $\lambda_t=1$ removes scaffold construction and yields the lowest CLIP score of $31.16$. We therefore use the scheduled $\lambda_t$ shown in~\Cref{fig:lambda_graph}, which applies stronger scaffold alignment at early denoising steps and gradually relaxes it. This schedule balances global coherence and local detail, producing a coherent coarse scaffold $Y_0^{\texttt{c}}$ whose refined output $Y_0^{\texttt{r}}$ achieves the highest CLIP score of $32.85$.

\begin{figure}[t]
\vspace{-2mm}
\centering
\begin{minipage}[b]{0.35\linewidth}
    \centering
    \scriptsize
    {\texttt{Planning}}
\end{minipage}
\begin{minipage}[b]{0.2\linewidth}
    \centering
    \scriptsize
    {\texttt{Image}}
\end{minipage}
\begin{minipage}[b]{0.43\linewidth}
    \centering
    \scriptsize
    {\texttt{Video}}
\end{minipage}
\includegraphics[width=1.\textwidth,]{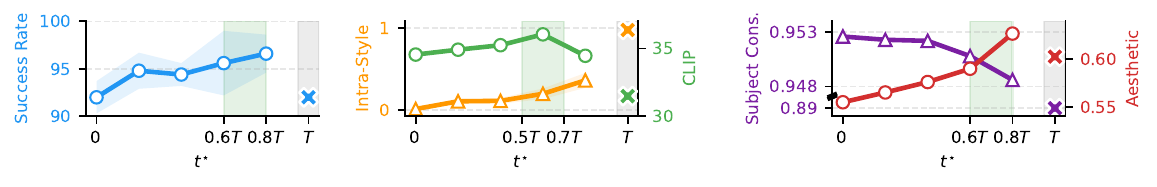}
\vspace{-6mm}
\caption{\textbf{Effect of refinement depth $t^\star$.}
We vary $t^\star$ to measure the trade-off between \purple{\textit{global structure preservation}} and \green{\textit{local detail recovery}}.
Image metrics compare Intra-Style (\textit{global}) with CLIP (\textit{local}), while video metrics compare subject consistency (\textit{global}) with frame-level aesthetic quality (\textit{local}). The colored $\mathbf{\times}$ denote generation from w/o scaffold alignment ($\ie$ $Y^{\texttt{r}}_{t^{\star}=T} \sim \calN(0, I_{\scriptsize{d}})$ in~\eqref{eq:structure_preserving_noise}) and intermediate $t^\star \in [0.5T,0.8T]$ values form a \hlc[bggreen]{\texttt{sweet spot}} across domains.}
\label{fig:fig_ablation_t_star} 
\vspace{-4mm}
\end{figure}

We next examine the refinement depth $t^\star$ in~\eqref{eq:structure_preserving_noise}, which controls the strength of the scaffold constraint during the second stage. Smaller $t^\star$ preserves the  global structure $Y_0^{\texttt{c}}$ but limits the the refinement stage to correct local artifacts. Larger $t^\star$ allows the local prior to modify local details more substantially, but increases the risk that the sample deviates from the global scaffold $Y_0^{\texttt{c}}$. As shown in~\Cref{fig:fig_ablation_t_star}, both extremes are suboptimal. At $t^\star=0$ (\ie w/o refinement in~\Cref{alg:diffusion_refinement}), coarse-stage artifacts persist, showing degraded local quality (lower CLIP and aesthetic scores) despite strong global coherence, whereas at $t^\star=T$ (\ie w/o scaffold alignment in~\Cref{alg:coarse_plan}), the global scaffold is discarded entirely, showing poor global coherence (higher style loss and lower subject consistency). Intermediate values $t^\star \in [0.5T, 0.8T]$ consistently yield the best balance between global structure preservation and local detail recovery across all three domains.

\section{Conclusion}

We introduced CoFi, an inference-time sampler that decomposes compositional generation into a consensus-based coarse stage and a structure-preserving refinement stage. By separating global structure formation from local detail recovery, we improve both global coherence and local quality over existing compositional baselines across robotic planning, panoramic image generation, and long video generation, while requiring $2$--$8\times$ fewer denoiser evaluations.

\paragraph{Limitations.}
The quality of the composed output is bounded by the expressiveness of the pretrained local prior, and the consensus schedule $\lambda_t$ and refinement depth $t^\star$ require per-domain selection. Moreover, the current consensus applies uniform weighting to all local plans, which may be suboptimal when segments differ in structural importance.

\clearpage

\bibliographystyle{assets/plainnat}
\bibliography{main}

\appendix
\onecolumn

\section{Derivation of the Closed-Form Coarse Estimate}\label{app:derivation}

We derive the closed-form solution of the coarse alignment objective in~\eqref{eq:consensus minimization}. For notational clarity, we drop the time index $t$ and write $\hat{X}^i := \hat{X}^i_{0|t}$ for the Tweedie estimate of the $i$-th local plan. Let $\gamma > 0$ denote the regularization strength and define
\begin{equation}\label{eq:consensus_obj_app}
f(Z^1, \dots, Z^N)
\;=\;
\sum_{i=1}^{N}
\left[
\|Z^i - \hat{X}^i\|^2
\;+\; \gamma \,\|Z^i - \bar{Z}\|^2
\right],
\qquad
\bar{Z} := \tfrac{1}{N}\textstyle\sum_{i=1}^{N} Z^i.
\end{equation}

\begin{proof}[Proof of~\eqref{eq:shared consensus}]
Since~\eqref{eq:consensus_obj_app} is a sum of convex quadratics in $(Z^1, \dots, Z^N)$, the objective is strictly convex and admits a unique global minimizer. We derive the first-order optimality conditions below.

Let $\delta_{ij}$ denote the Kronecker delta. Differentiating~\eqref{eq:consensus_obj_app} with respect to $Z^j$ and applying the chain rule with $\partial \bar{Z} / \partial Z^j = \frac{1}{N} I_m$ yields
\begin{equation}\label{eq:consensus_grad}
\frac{1}{2}\frac{\partial f}{\partial Z^j}
\;=\;
\bigl(Z^j - \hat{X}^j\bigr)
\;+\; \gamma \sum_{i=1}^{N} \bigl(Z^i - \bar{Z}\bigr) \Bigl(\delta_{ij} - \tfrac{1}{N}\Bigr).
\end{equation}
By construction, $\sum_{i=1}^{N}(Z^i - \bar{Z}) = \mathbf{0}$. Expanding the summation in~\eqref{eq:consensus_grad} and applying this identity, we obtain
\begin{equation}\label{eq:consensus_simplified}
\frac{1}{2}\frac{\partial f}{\partial Z^j}
\;=\;
\bigl(Z^j - \hat{X}^j\bigr)
\;+\; \gamma \bigl(Z^j - \bar{Z}\bigr).
\end{equation}
Setting~\eqref{eq:consensus_simplified} to zero and solving for $Z^j$:
\begin{equation}\label{eq:consensus_foc}
(1 + \gamma)\,Z^j \;=\; \hat{X}^j + \gamma\,\bar{Z}
\qquad \Longrightarrow \qquad
Z^j \;=\; \lambda\, \hat{X}^j \;+\; (1-\lambda)\,\bar{Z},
\quad \text{where } \lambda := \tfrac{1}{1+\gamma}.
\end{equation}
Summing both sides of~\eqref{eq:consensus_foc} over $j = 1, \dots, N$ and dividing by $N$ gives $\bar{Z} = \lambda\, \bar{X} + (1-\lambda)\,\bar{Z}$, where $\bar{X} := \frac{1}{N}\sum_{i=1}^{N}\hat{X}^i$. Rearranging yields $\lambda\,\bar{Z} = \lambda\,\bar{X}$, and hence
\begin{equation}\label{eq:consensus_mean_result}
\bar{Z}^\star \;=\; \bar{X}.
\end{equation}
Substituting~\eqref{eq:consensus_mean_result} into~\eqref{eq:consensus_foc} produces the minimizer $Z^{j,\star} = \lambda\, \hat{X}^j + (1-\lambda)\,\bar{X}$, which coincides with~\eqref{eq:shared consensus} upon restoring the time index $t$ and setting $\lambda_t := 1/(1+\gamma_t)$.
\end{proof}

\section{Experimental Details}\label{app:exp_details}

We adopt the experimental protocol of \citet{mishra2026compositional} and use the same pretrained diffusion models, datasets, and evaluation metrics. Below we provide the hyperparameter settings specific to our method.

\subsection{Compositional Long-Horizon Robotic Planning}\label{app:planning}

\paragraph{Task description.}
We evaluate on OGBench~\citep{park2025ogbench}, which provides offline datasets for goal-conditioned robotic tasks. The agent must navigate from a given start state to a goal state, and the offline dataset contains only short trajectory segments that do not individually solve the evaluation tasks. Successful planning therefore requires composing multiple short segments into a coherent long-horizon plan that reaches the goal while respecting environment constraints.

\begin{figure}[h]
\centering
{\scriptsize \texttt{PointMaze-M} ($N=3$)}\\[2pt]
\includegraphics[width=0.19\textwidth]{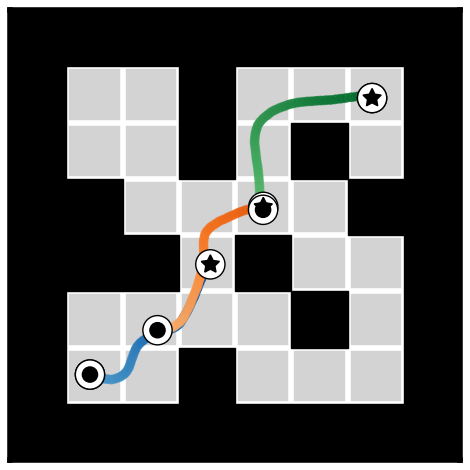}\hfill
\includegraphics[width=0.19\textwidth]{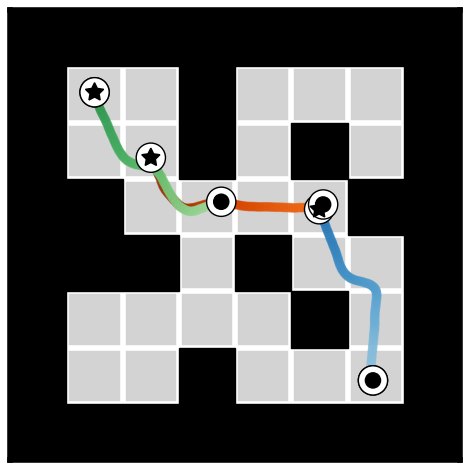}\hfill
\includegraphics[width=0.19\textwidth]{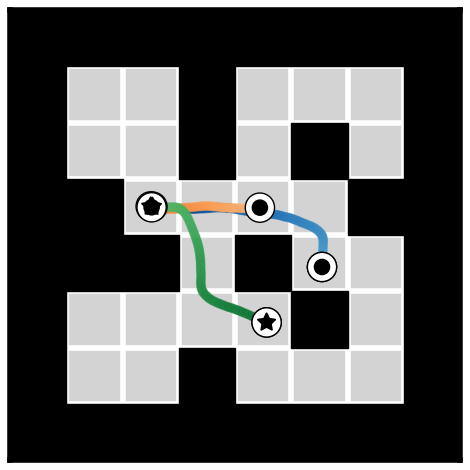}\hfill
\includegraphics[width=0.19\textwidth]{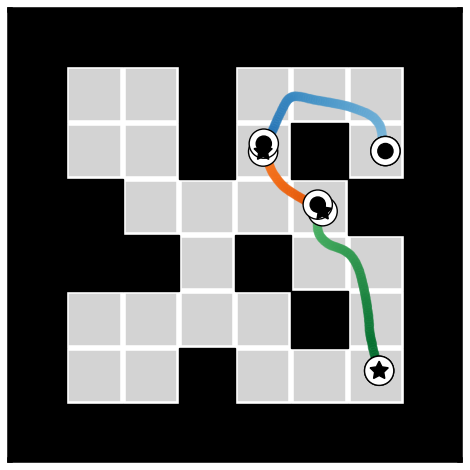}\hfill
\includegraphics[width=0.19\textwidth]{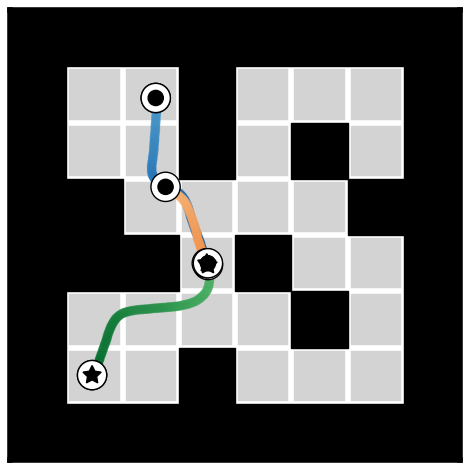}\\[4pt]
{\scriptsize \texttt{PointMaze-L} ($N=6$)}\\[2pt]
\includegraphics[width=0.19\textwidth]{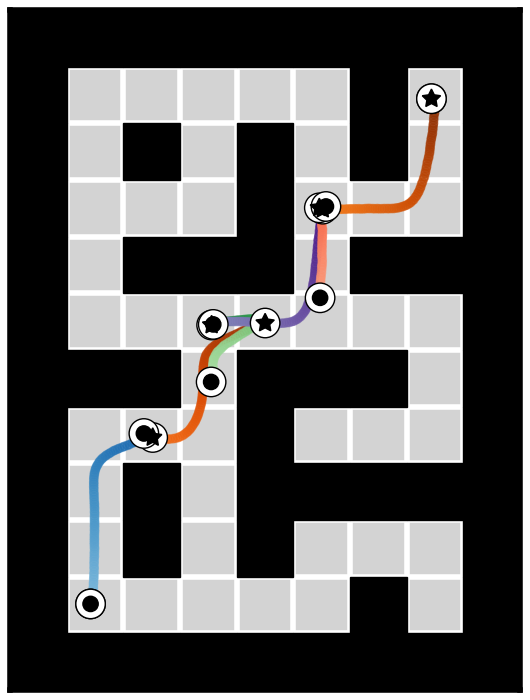}\hfill
\includegraphics[width=0.19\textwidth]{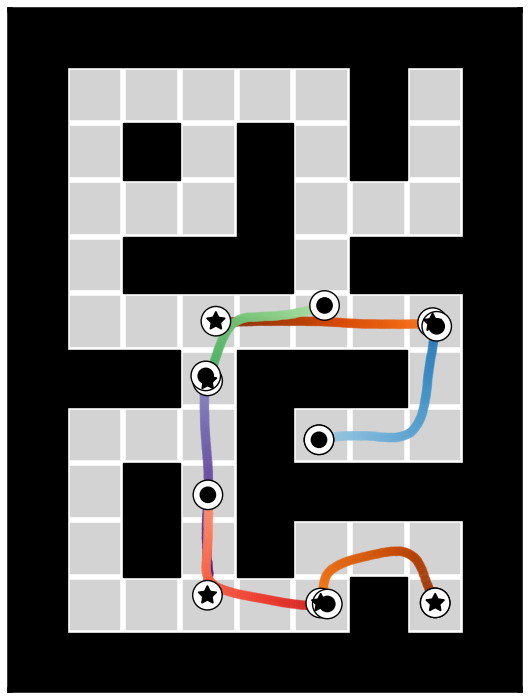}\hfill
\includegraphics[width=0.19\textwidth]{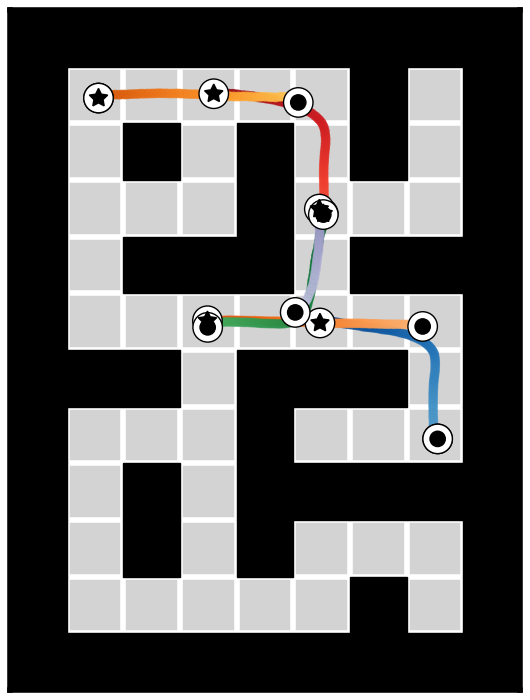}\hfill
\includegraphics[width=0.19\textwidth]{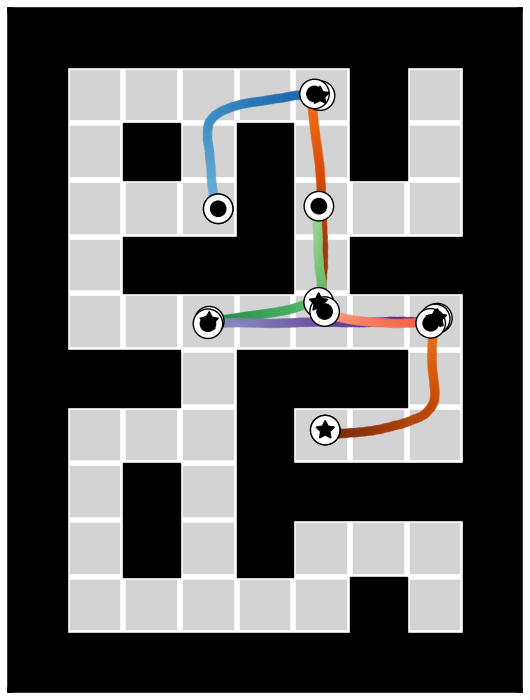}\hfill
\includegraphics[width=0.19\textwidth]{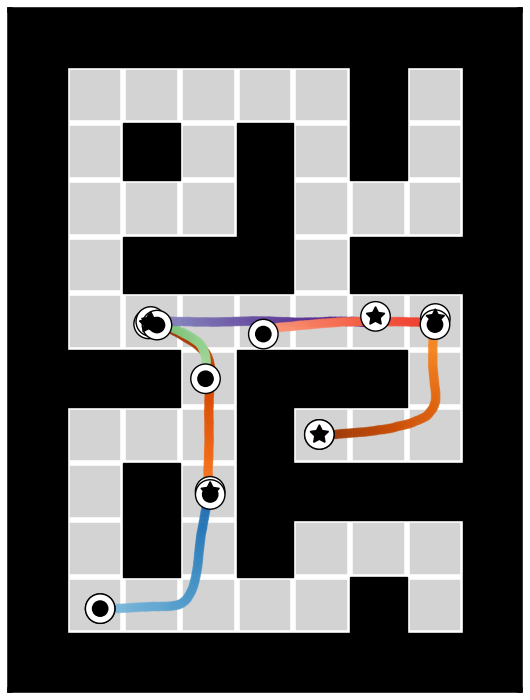}\\[4pt]
{\scriptsize \texttt{PointMaze-G} ($N=8$)}\\[2pt]
\includegraphics[width=0.19\textwidth]{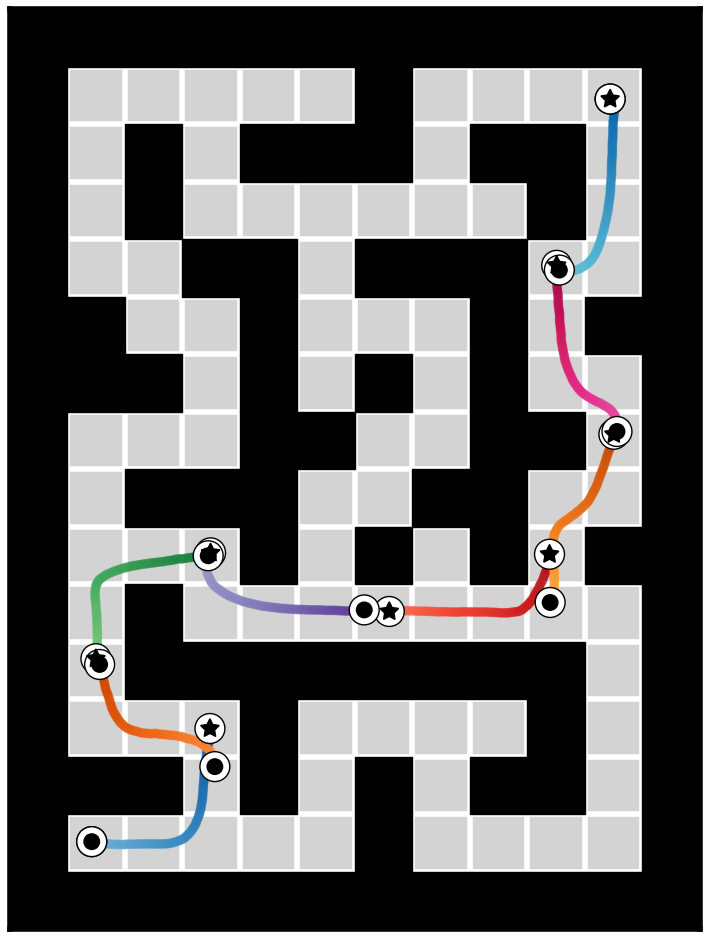}\hfill
\includegraphics[width=0.19\textwidth]{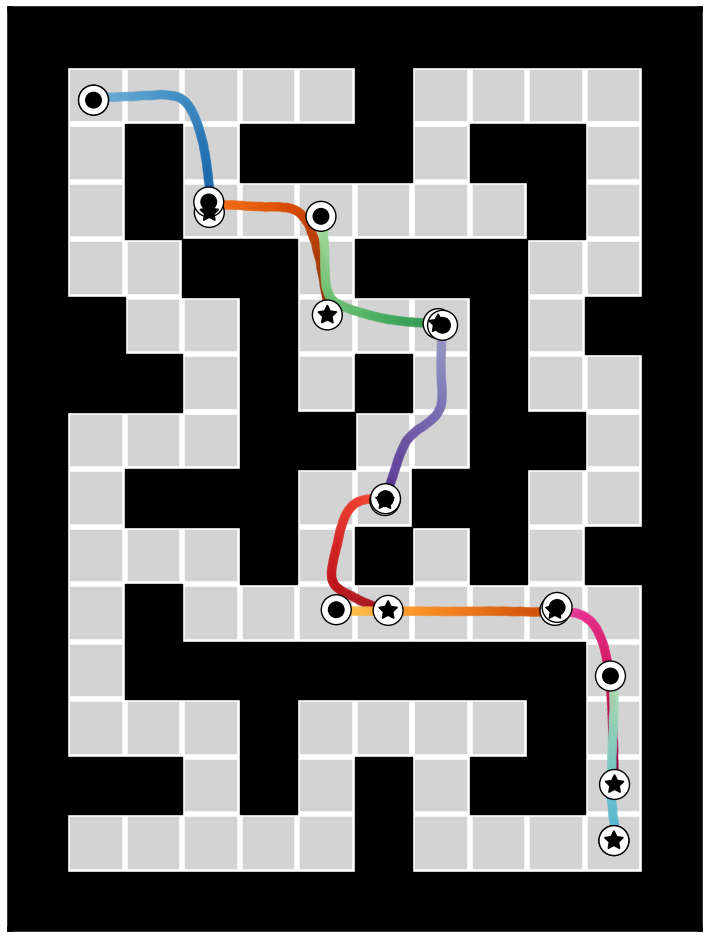}\hfill
\includegraphics[width=0.19\textwidth]{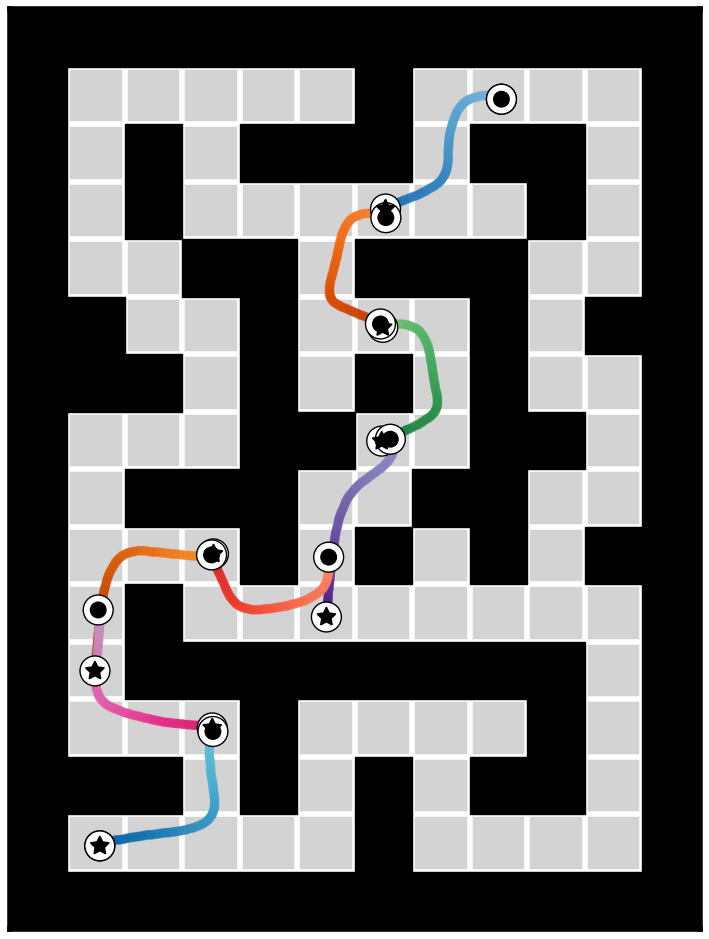}\hfill
\includegraphics[width=0.19\textwidth]{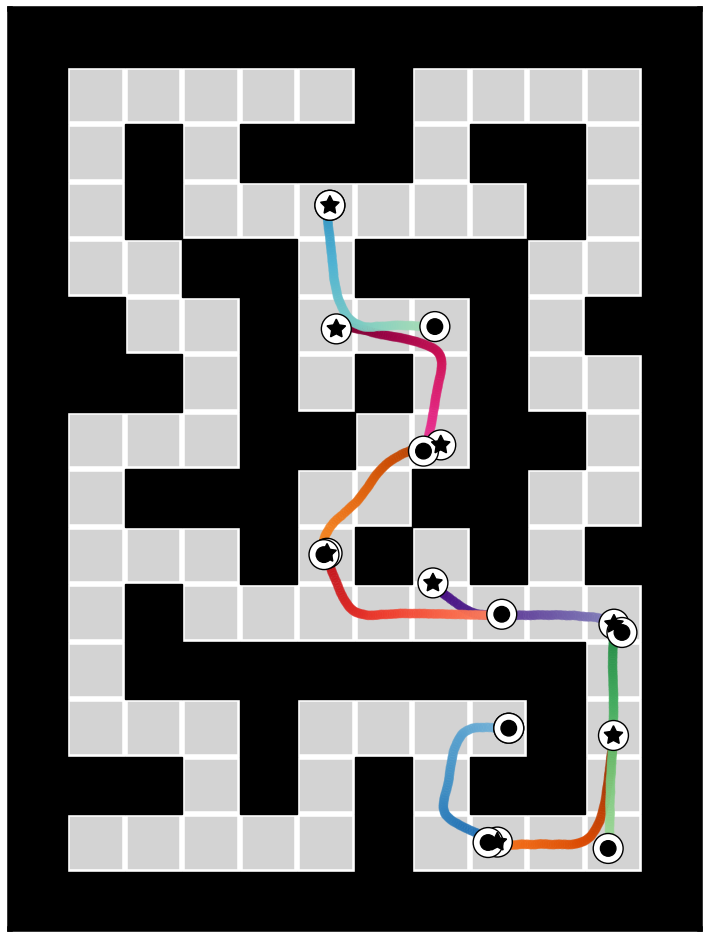}\hfill
\includegraphics[width=0.19\textwidth]{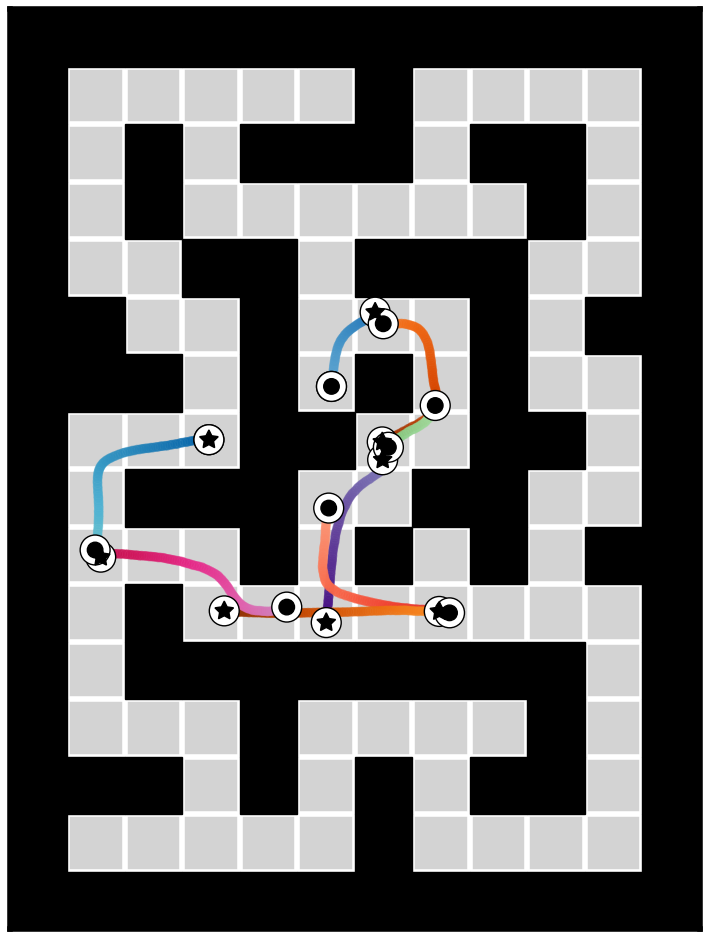}
\caption{\textbf{PointMaze trajectory examples.} Each row shows five evaluation tasks at increasing maze scale. Each color denotes a distinct local plan segment.}
\label{fig:pointmaze_all}
\end{figure}

\begin{itemize}[leftmargin=10pt]
\item \texttt{PointMaze} (\texttt{Stitch}): A 2D point agent navigates through a maze from a start position to a goal position. The observation and goal spaces are both 2D (XY position). We evaluate on \texttt{Medium}, \texttt{Large}, and \texttt{Giant} map sizes, where increasing scale requires more composed segments and longer navigation paths through the maze.

\item \texttt{AntMaze} (\texttt{Stitch}): An 8-DoF ant robot navigates through the same maze layouts as \texttt{PointMaze}. Although the full observation is 29-dimensional, planning is performed in the 2D XY subspace, and a learned inverse-dynamics model maps the planned waypoints back to the full action space for execution.

\item \texttt{Scene} (\texttt{Play}): A tabletop manipulation task with a robot gripper and multiple objects. The planning space is 14-dimensional, selected from the full 40-dimensional observation: gripper 3D position (indices 12--14), two object 3D positions (indices 19--21, 26--29), object orientations (indices 32--33, 36, 38), and contact states (indices 8--11). The environment defines 5 predefined tasks, each specifying a distinct start and goal configuration that requires the gripper to reach, grasp, and reposition one or more objects on the table. Unlike the maze tasks where the agent must find a collision-free path through a static environment, \texttt{Scene-Play} requires coordinating gripper motion with object interaction, and the offline play dataset contains unstructured exploration rather than goal-directed demonstrations. We evaluate 20 trials per task for a total of 100 episodes per seed. During execution, contact dimensions (indices 8--11) are snapped to binary values $\{0, 1\}$ to match the discrete nature of contact events in the training data.
\end{itemize}

Table~\ref{tab:env_specs} summarizes the per-environment configuration, including the observation and goal dimensions, local plan horizon and overlap length, default number of composed segments $N$, replanning distance threshold, and maximum environment steps.

\begin{table}[h]
\caption{Environment specifications for robotic planning.}
\centering
\renewcommand{\arraystretch}{1.1}
\setlength{\tabcolsep}{4pt}
\begin{tabular}{@{} l c c c c c c c @{}}
\toprule
Environment & obs dim & goal dim & horizon & overlap & $N$ & replan thres & max steps \\
\midrule
\texttt{PointMaze-Medium}  & 2  & 2  & 160 & 56 & 3 & 1.0 & 1000 \\
\texttt{PointMaze-Large}   & 2  & 2  & 160 & 56 & 6 & 1.0 & 1000 \\
\texttt{PointMaze-Giant}   & 2  & 2  & 160 & 56 & 8 & 1.0 & 1000 \\
\texttt{AntMaze-Medium}    & 29 & 2  & 160 & 56 & 3 & 4.0 & 1000 \\
\texttt{AntMaze-Large}     & 29 & 2  & 160 & 56 & 6 & 4.0 & 1000 \\
\texttt{AntMaze-Giant}     & 29 & 2  & 160 & 56 & 9 & 4.0 & 2000 \\
\texttt{Scene-Play}        & 40 & 14 & 128 & 48 & 8 & 2.0 & 1500 \\
\bottomrule
\end{tabular}
\label{tab:env_specs}
\end{table}

\paragraph{Guidance cost.}
The local guidance cost in~\eqref{eq:guided_DDIM} is the overlap mean-squared error between the predicted clean boundaries of adjacent segments:
\[
\hat{r}_t(X_t^i, X_t^{i+1})
= -c \left\| \texttt{R}(\hat{X}^i_{0|t}) - \texttt{L}(\hat{X}^{i+1}_{0|t}) \right\|^2,
\]
where $c = 5.0$. This gradient-based guidance is used in both the coarse stage (Algorithm~\ref{alg:coarse_plan}) and the refinement stage (Algorithm~\ref{alg:diffusion_refinement}). After each denoising step, the composed global trajectory is obtained by exponentially blending adjacent segments in their overlap regions~\citep{luo2025generative}. Concretely, for an overlap of length $L$, the blended trajectory at position $\ell \in \{0, \dots, L-1\}$ is
\begin{equation}\label{eq:exp_blend}
\texttt{Blend}(\ell) \;=\; w(\ell) \cdot \texttt{R}(X^i) \;+\; \bigl(1 - w(\ell)\bigr) \cdot \texttt{L}(X^{i+1}),
\qquad
w(\ell) \;=\; \frac{e^{-\beta \ell / (L-1)} - e^{-\beta}}{1 - e^{-\beta}},
\end{equation}
where $\beta = 2.0$ is used.

\paragraph{Diffusion model and hyperparameters.}
We use the same pretrained 1D U-Net diffusion models from \citet{luo2025generative, mishra2026compositional}, trained on the OGBench offline datasets with $512$ diffusion steps. Each local plan covers a trajectory segment of $160$ (maze) or $128$ (scene) timesteps, and $N$ overlapping segments are composed to cover the full start-to-goal trajectory. At inference time, we use $T = 100$ DDIM steps with stochasticity $\eta = 1.0$. For the coarse stage, we use a ramped $\lambda_t$ schedule: $\lambda_t$ starts at $\lambda_0 = 0.2$ and linearly increases to $1.0$. The refinement depth is set to $t^\star = 0.7T$ for \texttt{Maze}. We generate a batch of $40$ candidate trajectories per planning problem and select the one with the lowest overlap MSE.

\paragraph{Evaluation protocol.}
We report success rates over $100$ evaluation episodes averaged across $5$ random seeds. Each evaluation episode samples a random start-goal pair from the OGBench evaluation set. An episode is considered successful if the agent reaches within a threshold distance of the goal within the maximum number of environment steps (see~Table~\ref{tab:env_specs}). For non-compositional baselines (GCBC, GCIVL, GCIQL, HIQL), we use the results reported in \citet{mishra2026compositional}.

\paragraph{Replanning and execution.}
For maze tasks, we follow \citet{luo2025generative} and use distance-based replanning: the agent replans when it deviates beyond the threshold distance (see~Table~\ref{tab:env_specs}) from the current planned trajectory. Actions are extracted from the planned waypoints using a learned inverse-dynamics model that maps consecutive observation pairs to low-level actions. For \texttt{Scene-Play}, the agent replans up to $10$ times with a distance threshold of $2.0$, and the planned trajectory is executed for up to $1500$ environment steps per episode.

\begin{figure}[h]
\centering
{\scriptsize \texttt{AntMaze-M} ($N=3$)}\\[2pt]
\includegraphics[width=0.19\textwidth]{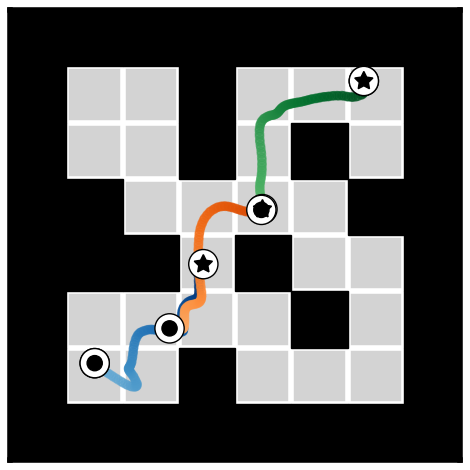}\hfill
\includegraphics[width=0.19\textwidth]{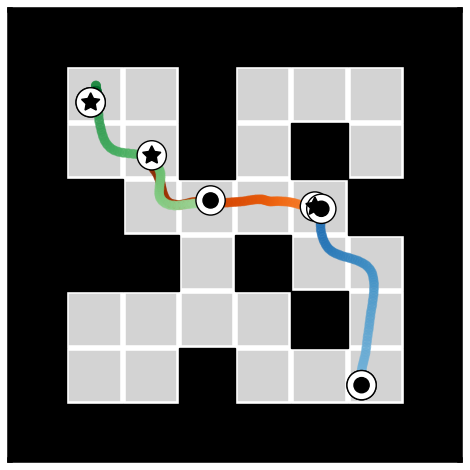}\hfill
\includegraphics[width=0.19\textwidth]{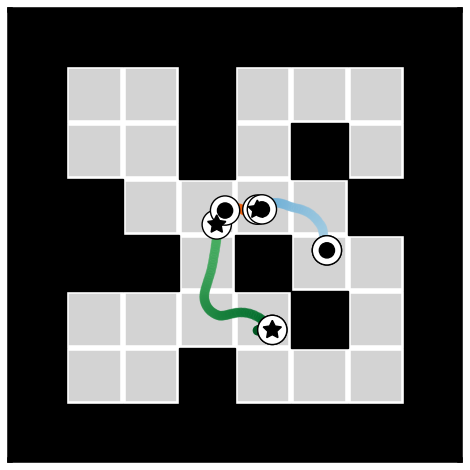}\hfill
\includegraphics[width=0.19\textwidth]{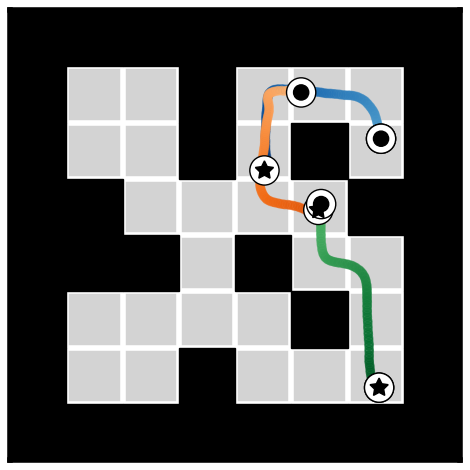}\hfill
\includegraphics[width=0.19\textwidth]{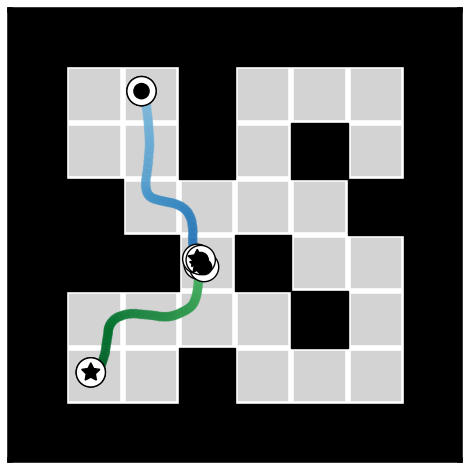}\\[4pt]
{\scriptsize \texttt{AntMaze-L} ($N=6$)}\\[2pt]
\includegraphics[width=0.19\textwidth]{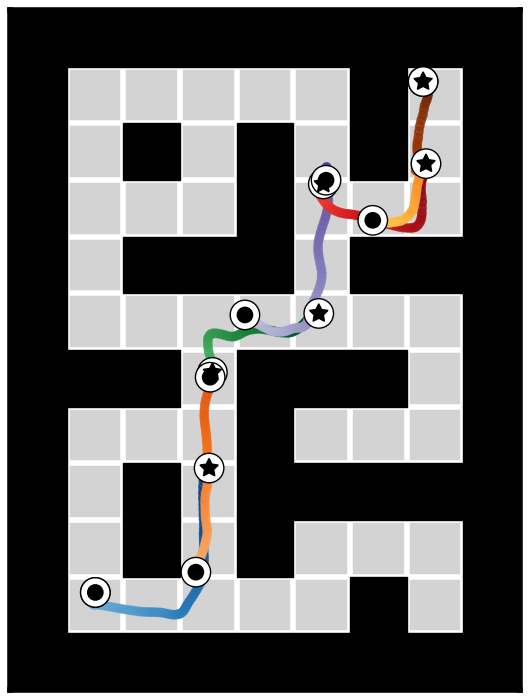}\hfill
\includegraphics[width=0.19\textwidth]{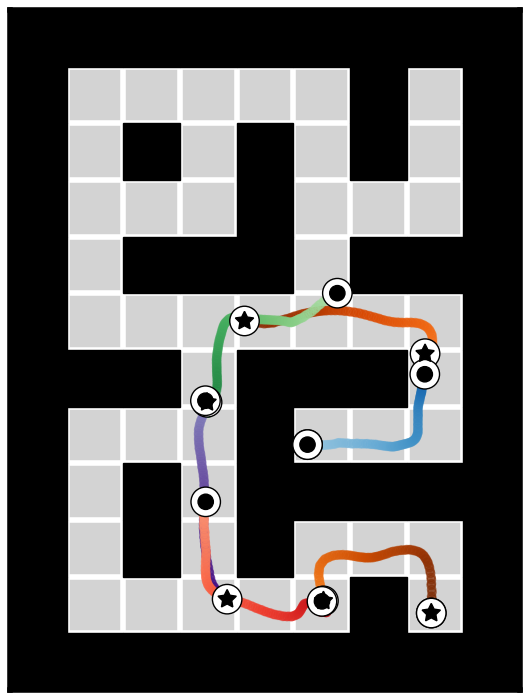}\hfill
\includegraphics[width=0.19\textwidth]{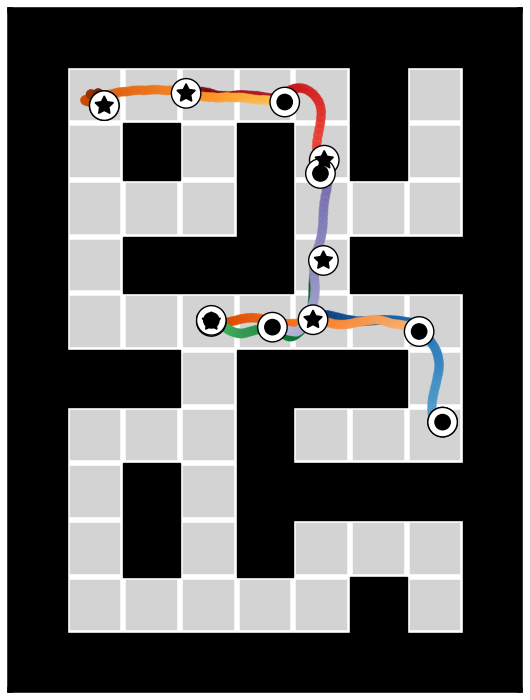}\hfill
\includegraphics[width=0.19\textwidth]{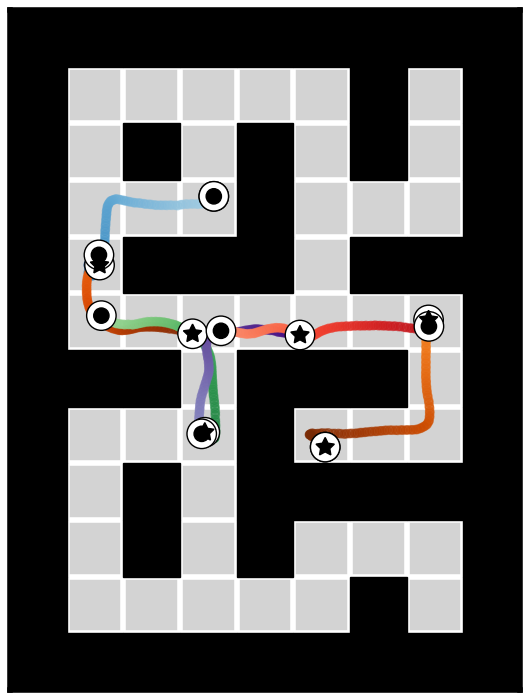}\hfill
\includegraphics[width=0.19\textwidth]{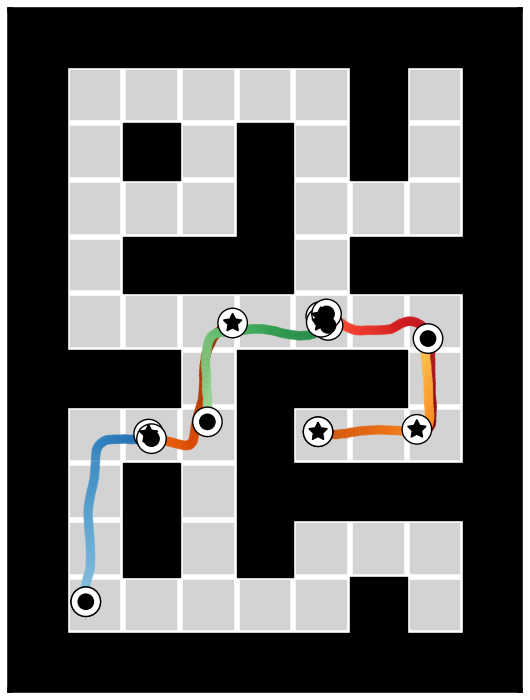}\\[4pt]
{\scriptsize \texttt{AntMaze-G} ($N=9$)}\\[2pt]
\includegraphics[width=0.19\textwidth]{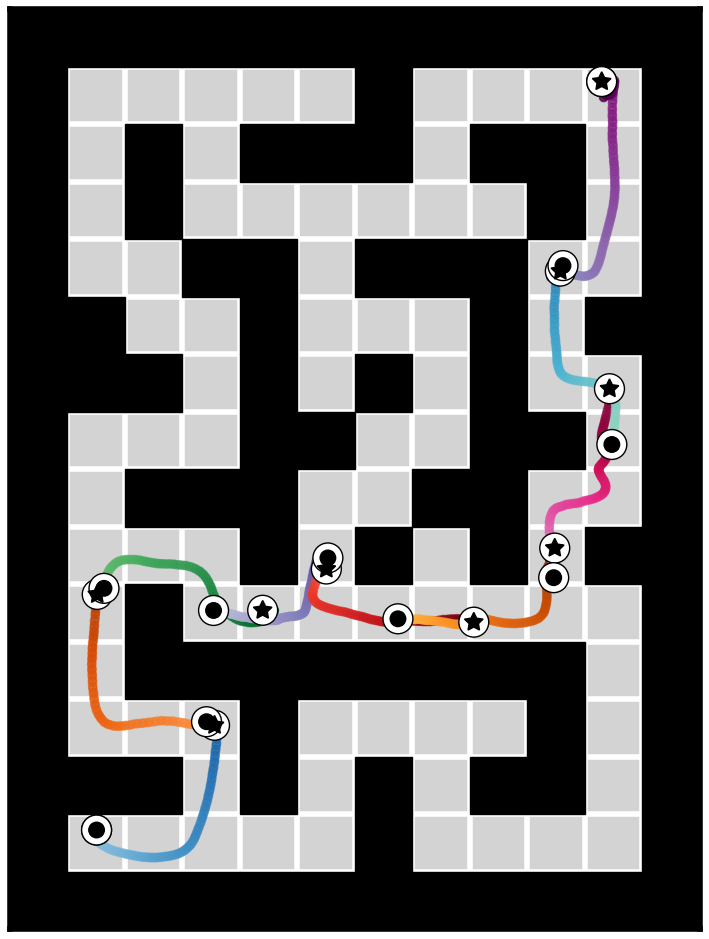}\hfill
\includegraphics[width=0.19\textwidth]{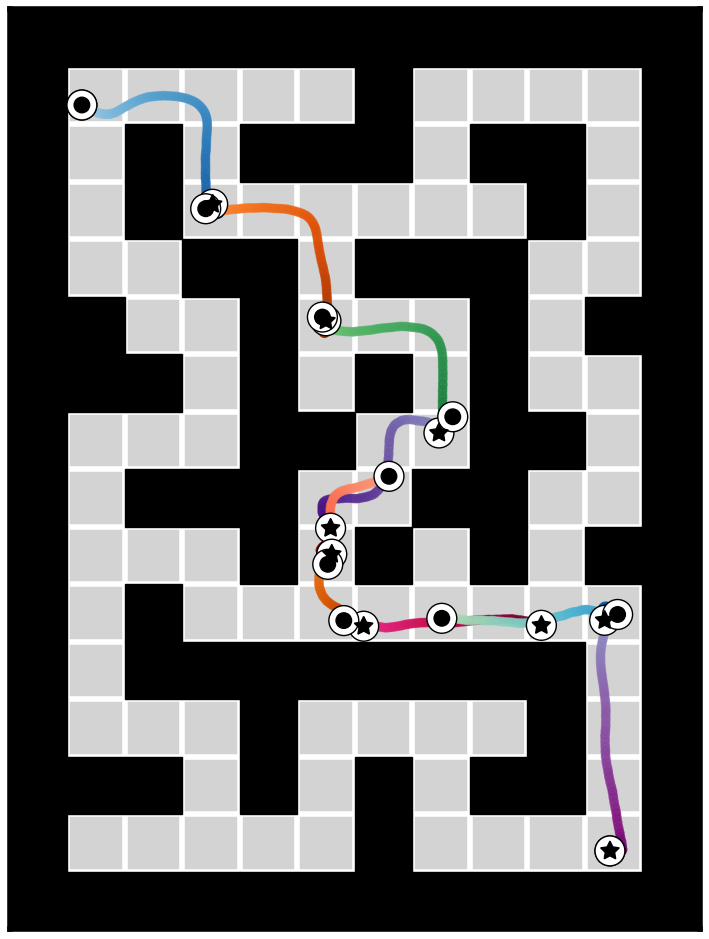}\hfill
\includegraphics[width=0.19\textwidth]{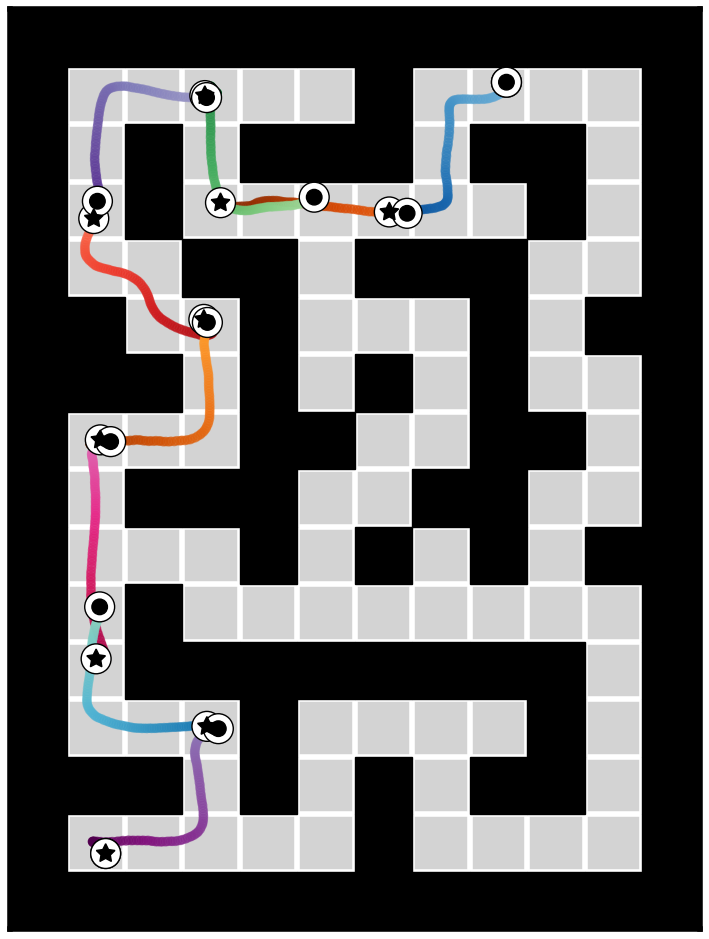}\hfill
\includegraphics[width=0.19\textwidth]{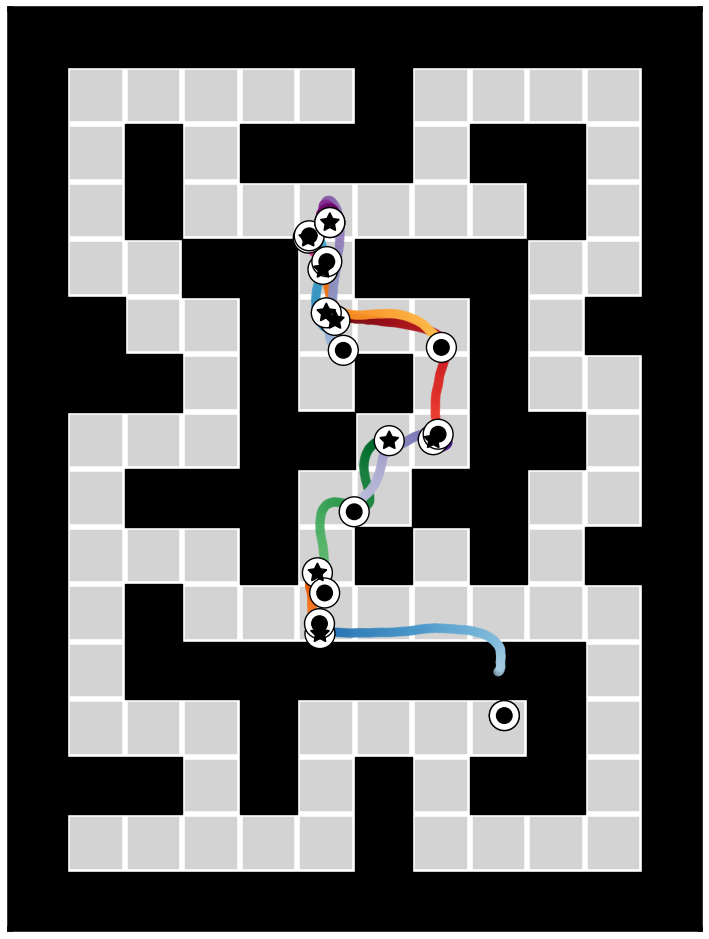}\hfill
\includegraphics[width=0.19\textwidth]{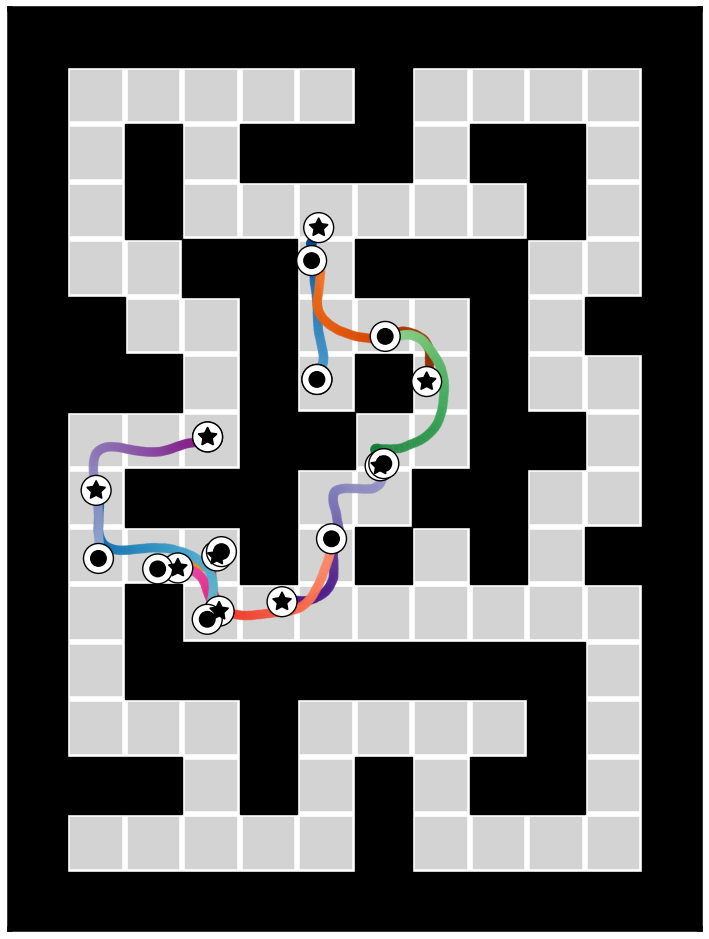}
\caption{\textbf{AntMaze trajectory examples.} Each row shows five evaluation tasks at increasing maze scale. Each color denotes a distinct local plan segment.}
\label{fig:antmaze_all}
\end{figure}

\begin{figure}[h]
\centering
{\scriptsize \texttt{Start} \hfill \texttt{Goal}}\\[1pt]
\includegraphics[width=\textwidth,height=1.8cm,keepaspectratio=false]{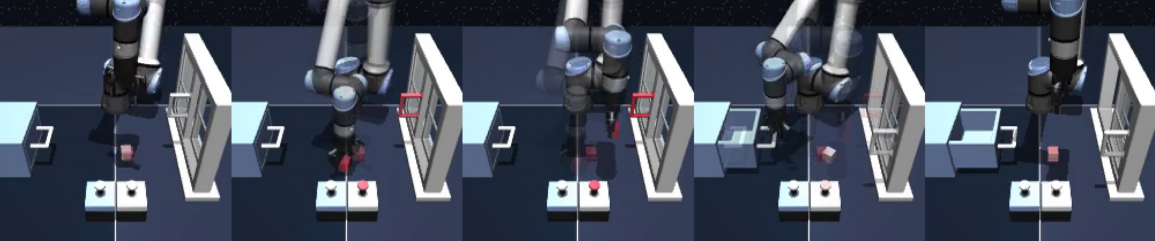}\\[-1pt]
{\tiny \color{gray!80!black}\texttt{Task 1}}\\[2pt]
\includegraphics[width=\textwidth,height=1.8cm,keepaspectratio=false]{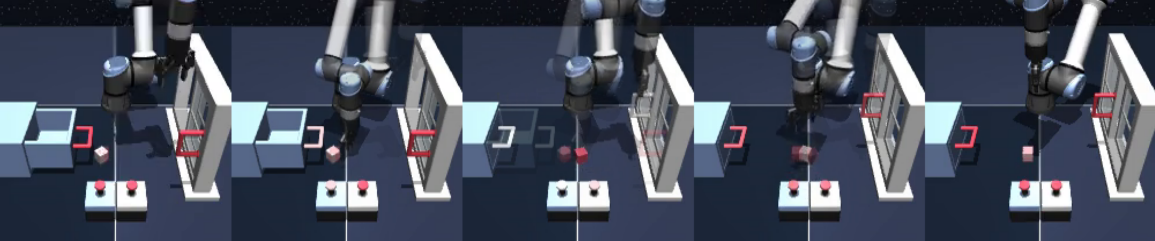}\\[-1pt]
{\tiny \color{gray!80!black}\texttt{Task 2}}\\[2pt]
\includegraphics[width=\textwidth,height=1.8cm,keepaspectratio=false]{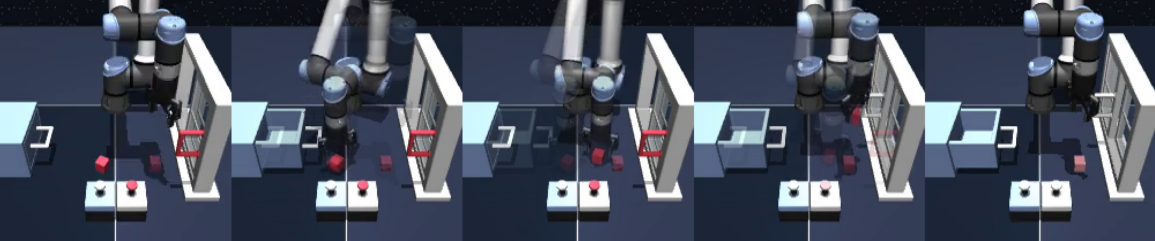}\\[-1pt]
{\tiny \color{gray!80!black}\texttt{Task 3}}\\[2pt]
\includegraphics[width=\textwidth,height=1.8cm,keepaspectratio=false]{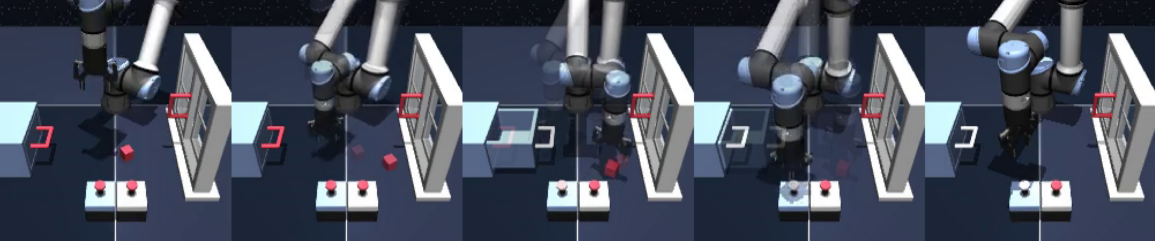}\\[-1pt]
{\tiny \color{gray!80!black}\texttt{Task 4}}\\[2pt]
\includegraphics[width=\textwidth,height=1.8cm,keepaspectratio=false]{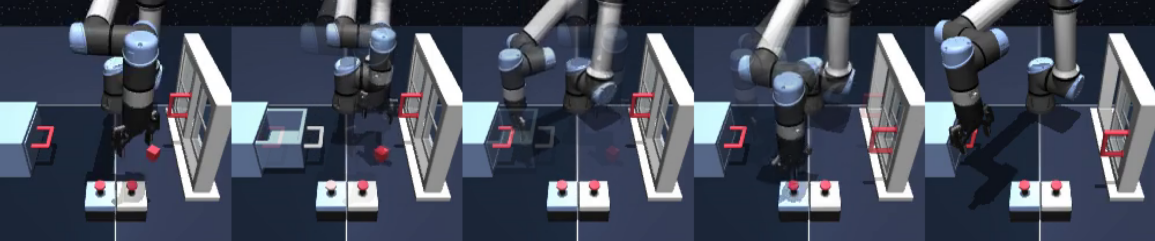}\\[-1pt]
{\tiny \color{gray!80!black}\texttt{Task 5}}
\caption{\textbf{Scene-Play rollout examples.} Each row shows keyframes from a successful rollout for one of the five predefined manipulation tasks. The robot gripper must reach, grasp, and reposition objects on the table according to the task-specific goal configuration.}
\label{fig:sceneplay_all}
\end{figure}

\paragraph{Per-task trajectory examples.}
Figures~\ref{fig:pointmaze_all} and~\ref{fig:antmaze_all} show composed trajectories generated by our method on each maze environment across different evaluation tasks, and Figure~\ref{fig:sceneplay_all} shows rollout keyframes for the five Scene-Play manipulation tasks. Each color denotes a distinct local plan segment, with overlapping boundaries between adjacent segments. The start position is marked with $\circ$ and the goal with $\star$. As the maze scale increases from \texttt{Medium} to \texttt{Giant}, the number of composed segments $N$ grows accordingly, and the coarse stage becomes increasingly important for maintaining a globally coherent route.

\subsection{Bethe Approximation for Compositional Diffusion}\label{app:bethe}

\paragraph{Factor graph formulation.}
Recall from~\Cref{sec:prelim} that the global plan $Y = [x^1, x^2, \dots, x^{2N+1}]$ is composed of $N$ overlapping local plans $X^i = [x^{2i-1}, x^{2i}, x^{2i+1}]$, where each adjacent pair $(X^i, X^{i+1})$ shares a boundary factor $x^{2i+1}$. This structure naturally defines a factor graph: each local plan $X^i$ corresponds to a \emph{factor node} with local prior $p_t(X^i_t)$, and each shared boundary $x^{2i+1}_t$ corresponds to a \emph{variable node} connecting the two adjacent factors. The joint distribution over the global plan factorizes over the factor graph as
\[
p_t(Y_t) \;\propto\; \prod_{i=1}^{N} p_t(X^i_t).
\]
Since the local plans overlap, this product double-counts the shared boundary variables. The Bethe free energy~\citep{yedidia2000generalized, yedidia2005constructing} corrects for this overcounting by subtracting the marginal log-densities at the shared boundaries:
\[\label{eq:bethe_free_energy}
\log p_t(Y_t)
\;\approx\;
\sum_{i=1}^{N} \log p_t(X^i_t)
\;-\;
\sum_{i=1}^{N-1} \log p_t(x^{2i+1}_t).
\]
The first sum aggregates the local log-densities (one per factor node), and the second sum removes the overcounted overlap marginals (one per shared variable node). This is the standard Bethe approximation applied to the chain-structured factor graph.

Differentiating~\eqref{eq:bethe_free_energy} with respect to the shared boundary $x^{2i+1}_t$ yields the compositional score:
\[
\nabla_{x^{2i+1}_t} \log p_t(Y_t)
\;\approx\;
\underbrace{\nabla \log p_t\!\left(\texttt{R}(X^i_t) \mid X^i_t\right)}_{\texttt{from left factor}}
\;+\;
\underbrace{\nabla \log p_t\!\left(\texttt{L}(X^{i+1}_t) \mid X^{i+1}_t\right)}_{\texttt{from right factor}}
\;-\;
\nabla \log p_t(x^{2i+1}_t),
\]
where $\texttt{R}(X^i_t) = \texttt{L}(X^{i+1}_t) = x^{2i+1}_t$ is the shared boundary. When the overlap marginal $p_t(x^{2i+1}_t)$ is approximated as uniform or absorbed into the local scores, this reduces to the form in~\eqref{eq:bethe approximation}: an average of the two conditional scores from the left and right factors.

\subsection{Panoramic Image Generation}\label{app:panorama}

\paragraph{Task description.}
Given a text prompt, the task is to generate a wide panoramic image ($512 \times 4608$) that maintains consistent style, color, and semantic content across the full spatial extent. The pretrained Stable Diffusion~2.0~\citep{rombach2022high} model generates $512 \times 512$ images natively, so producing a panorama requires composing $N = 9$ overlapping patches with $50\%$ spatial overlap.

\paragraph{Guidance.}
Local consistency across adjacent patches is enforced via the Bethe approximation in~\eqref{eq:bethe approximation}. Concretely, in the shared spatial overlap between patches $X^i$ and $X^{i+1}$, the fused denoised estimate is obtained by ramp blending rather than uniform averaging. At spatial position $\ell$ within an overlap of width $W$:
\begin{equation}\label{eq:ramp_blend_image}
\hat{X}^{\texttt{fused}}_{0|t}(\ell) \;=\; w(\ell) \cdot \hat{X}^i_{0|t} \;+\; \bigl(1 - w(\ell)\bigr) \cdot \hat{X}^{i+1}_{0|t},
\qquad
w(\ell) \;=\; 1 - \frac{\ell}{W},
\end{equation}
which linearly transitions from the left patch to the right. This ramp blending produces smoother boundary transitions than equal-weight averaging. Global coherence is achieved through the coarse mean regularization in~\eqref{eq:shared consensus}.

\paragraph{Hyperparameters.}
We use $T = 50$ DDIM steps with classifier-free guidance scale $7.5$. For the coarse stage, we set $\lambda_0 = 0.2$ with a linear ramp schedule that increases $\lambda_t$ to $1.0$. The clean estimate $\hat{X}^i_{0|t}$ is obtained from the fused (overlap-averaged) denoised output. We use sampled directional noise in the coarse stage, meaning each local plan receives independently sampled noise for the DDIM direction term rather than sharing a common noise realization. For the refinement stage, we set $t^\star = 0.7T$ (refinement begins from step $15$) with stochasticity $\eta = 1.0$ and ramp blending. We generate $5$ random seeds per prompt and report the mean and standard deviation across seeds.
\begin{figure}[t]
\centering
\begin{minipage}{0.49\textwidth}\centering\scriptsize\texttt{Coarse}\end{minipage}\hfill
\begin{minipage}{0.49\textwidth}\centering\scriptsize\texttt{Refinement}\end{minipage}\\[2pt]

\begin{minipage}{\textwidth}
\centering
\includegraphics[width=0.49\textwidth]{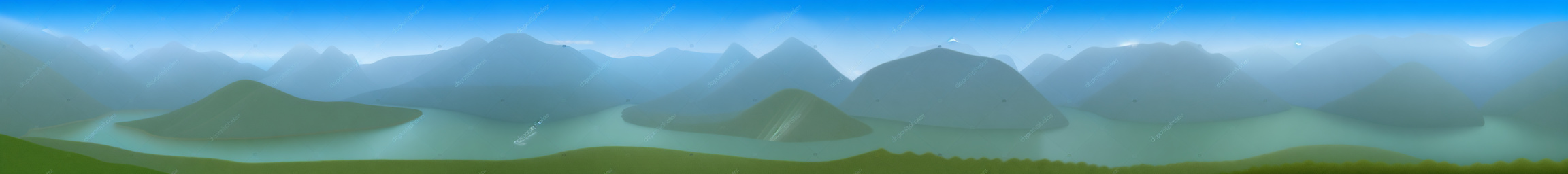}\hfill
\includegraphics[width=0.49\textwidth]{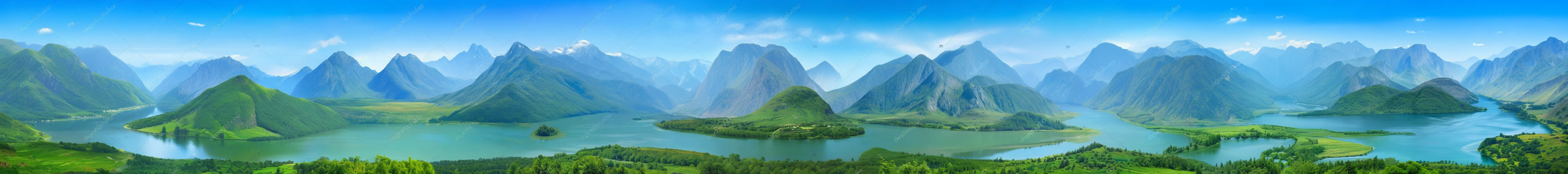}\\[-1pt]
{\tiny \color{gray!80!black}\texttt{landscape}}\\[3pt]
\end{minipage}

\begin{minipage}{\textwidth}
\centering
\includegraphics[width=0.49\textwidth]{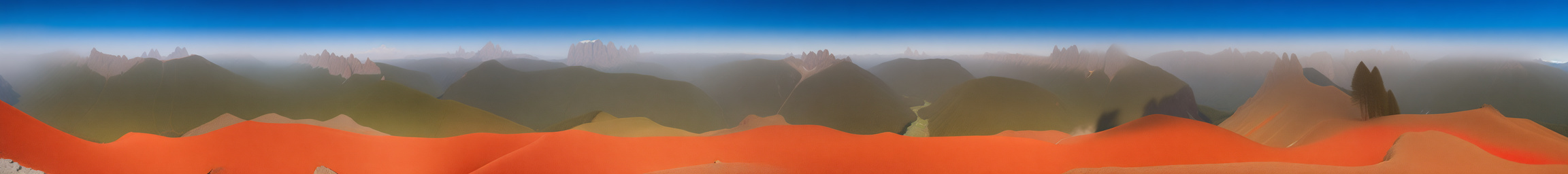}\hfill
\includegraphics[width=0.49\textwidth]{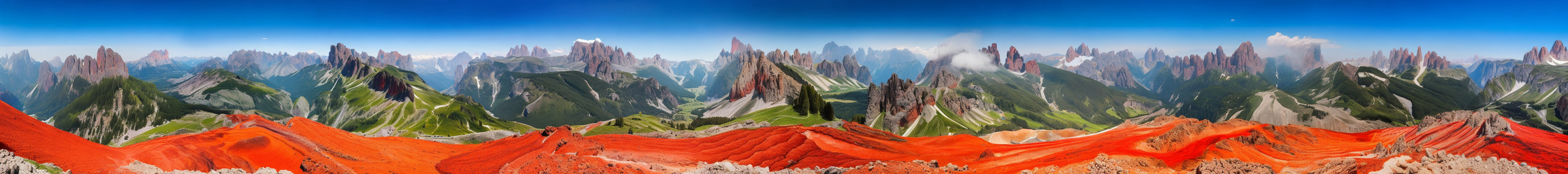}\\[-1pt]
{\tiny \color{gray!80!black}\texttt{dolomites}}\\[3pt]
\end{minipage}

\begin{minipage}{\textwidth}
\centering
\includegraphics[width=0.49\textwidth]{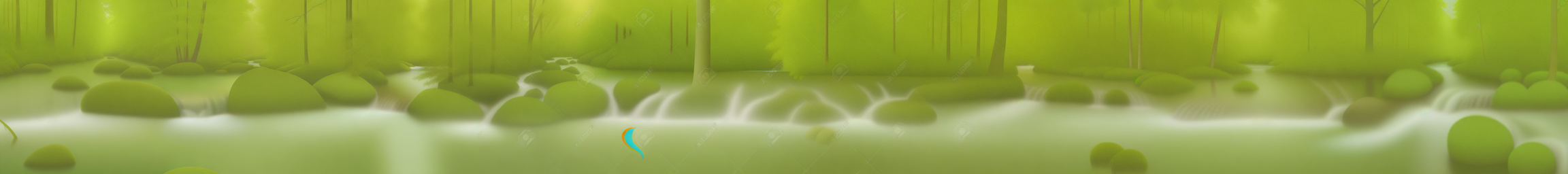}\hfill
\includegraphics[width=0.49\textwidth]{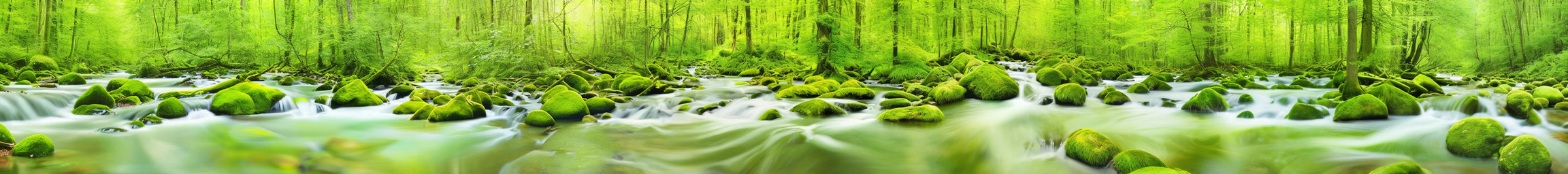}\\[-1pt]
{\tiny \color{gray!80!black}\texttt{babbling-brook}}\\[3pt]
\end{minipage}

\begin{minipage}{\textwidth}
\centering
\includegraphics[width=0.49\textwidth]{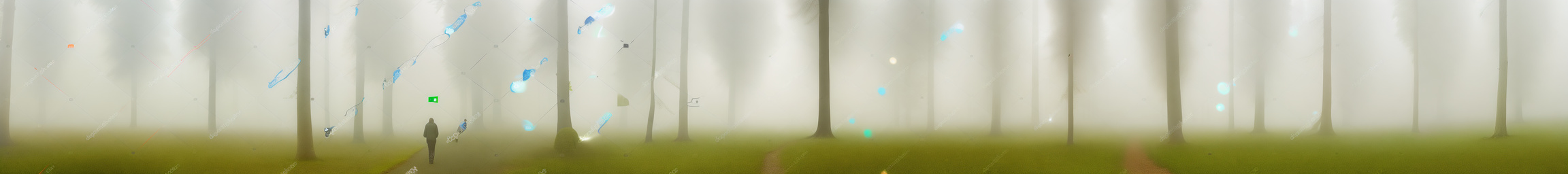}\hfill
\includegraphics[width=0.49\textwidth]{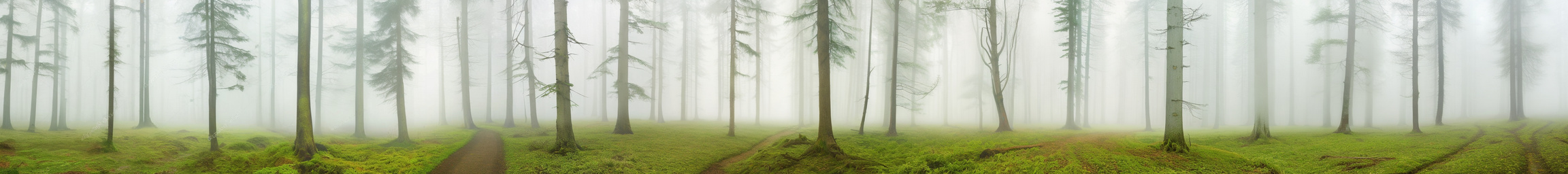}\\[-1pt]
{\tiny \color{gray!80!black}\texttt{misty-fog}}\\[3pt]
\end{minipage}

\begin{minipage}{\textwidth}
\centering
\includegraphics[width=0.49\textwidth]{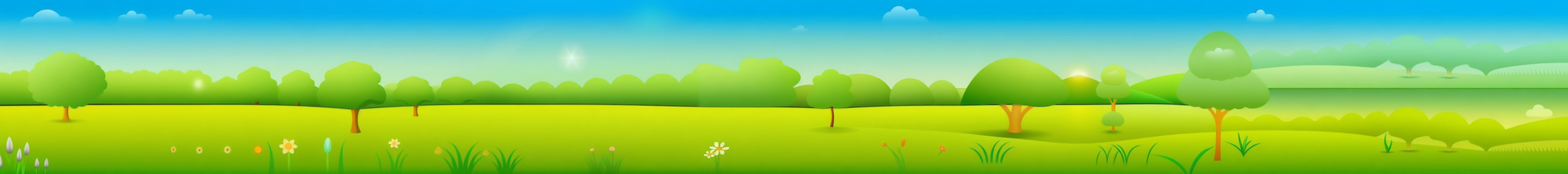}\hfill
\includegraphics[width=0.49\textwidth]{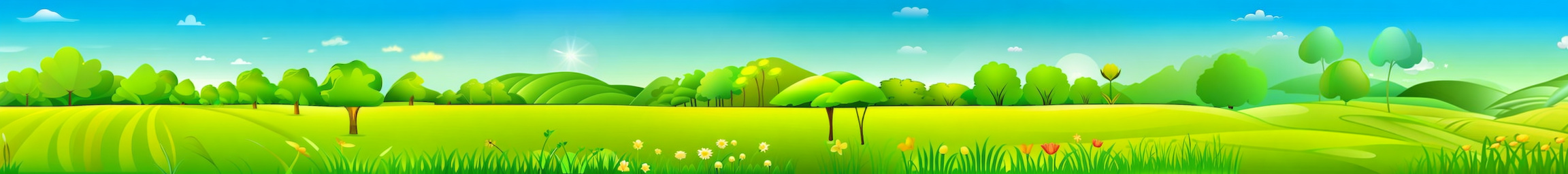}\\[-1pt]
{\tiny \color{gray!80!black}\texttt{cartoon}}\\[3pt]
\end{minipage}

\begin{minipage}{\textwidth}
\centering
\includegraphics[width=0.49\textwidth]{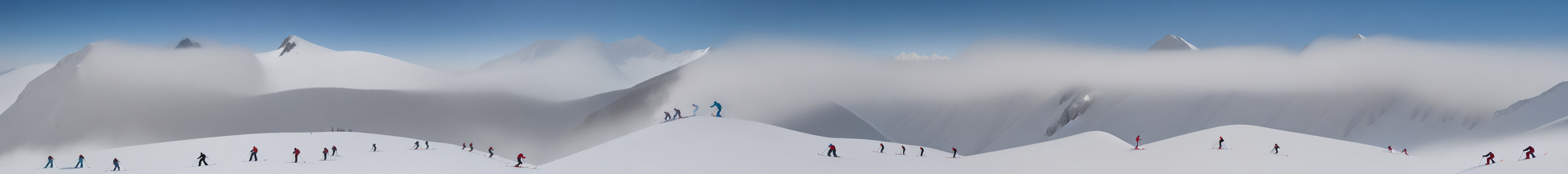}\hfill
\includegraphics[width=0.49\textwidth]{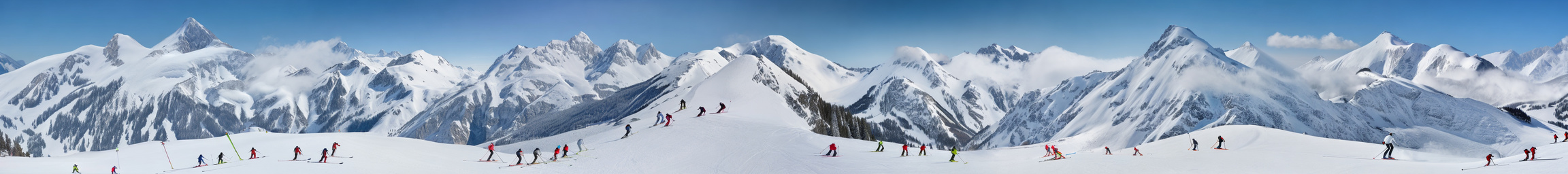}\\[-1pt]
{\tiny \color{gray!80!black}\texttt{skiers}}\\[3pt]
\end{minipage}

\begin{minipage}{\textwidth}
\centering
\includegraphics[width=0.49\textwidth]{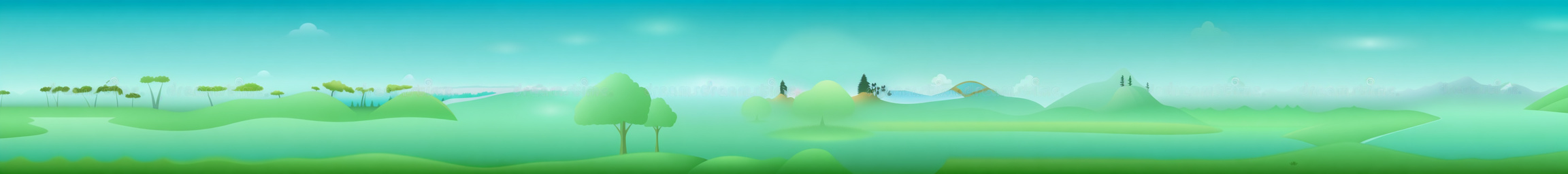}\hfill
\includegraphics[width=0.49\textwidth]{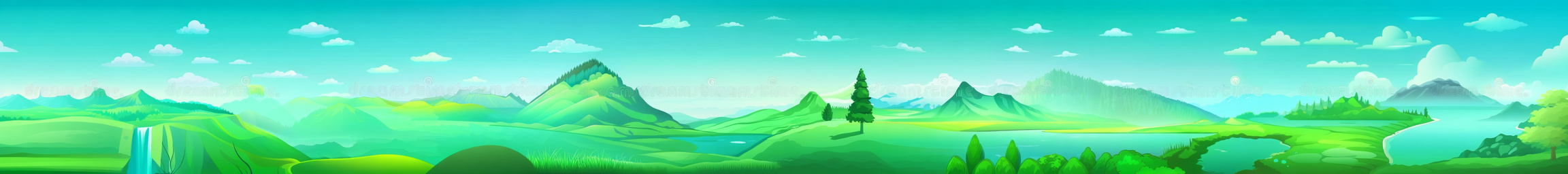}\\[-1pt]
{\tiny \color{gray!80!black}\texttt{anime}}\\[3pt]
\end{minipage}

\begin{minipage}{\textwidth}
\centering
\includegraphics[width=0.49\textwidth]{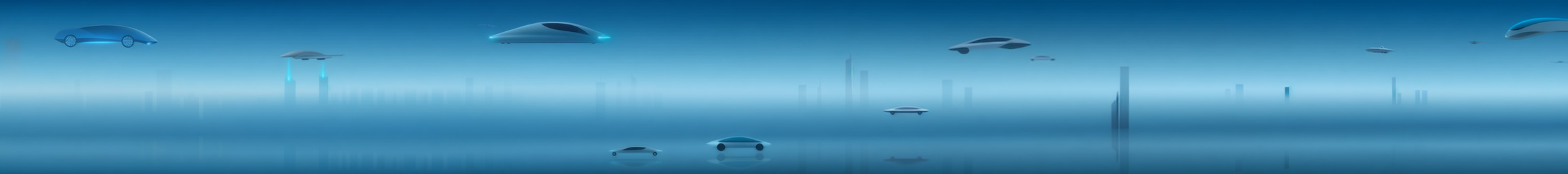}\hfill
\includegraphics[width=0.49\textwidth]{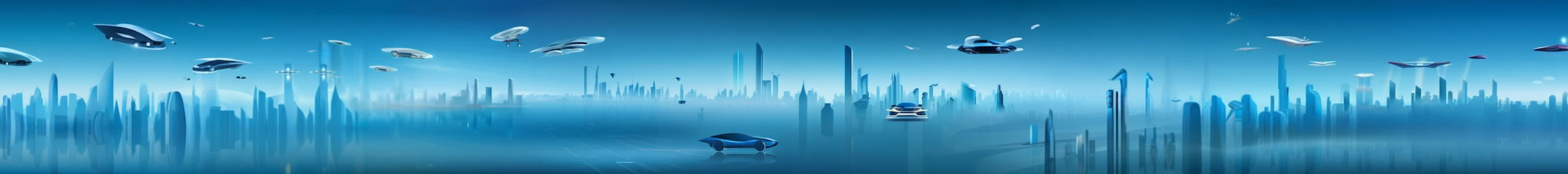}\\[-1pt]
{\tiny \color{gray!80!black}\texttt{futuristic}}\\[3pt]
\end{minipage}

\begin{minipage}{\textwidth}
\centering
\includegraphics[width=0.49\textwidth]{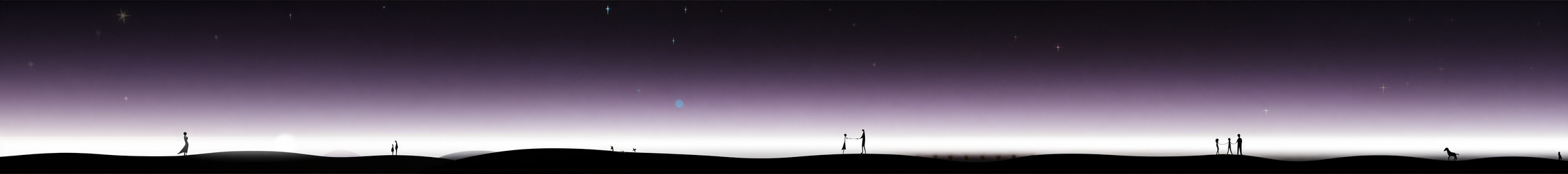}\hfill
\includegraphics[width=0.49\textwidth]{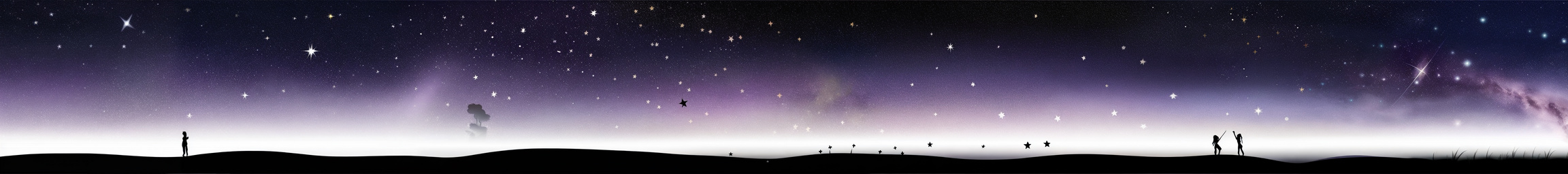}\\[-1pt]
{\tiny \color{gray!80!black}\texttt{silhouette}}\\[3pt]
\end{minipage}

\begin{minipage}{\textwidth}
\centering
\includegraphics[width=0.49\textwidth]{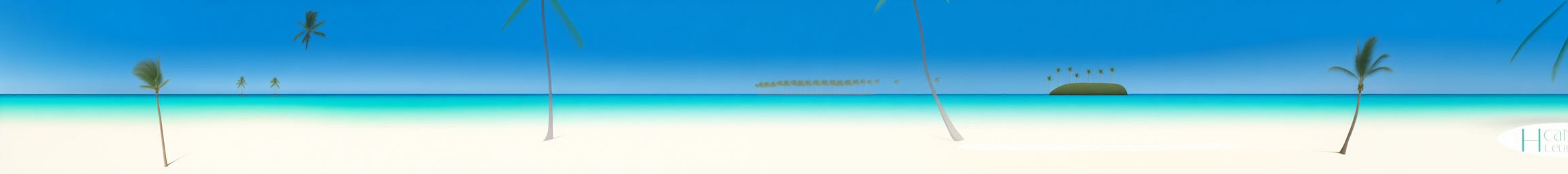}\hfill
\includegraphics[width=0.49\textwidth]{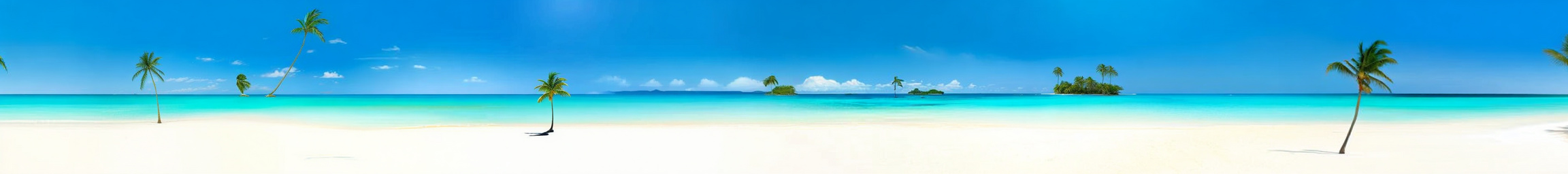}\\[-1pt]
{\tiny \color{gray!80!black}\texttt{palm-trees}}\\[3pt]
\end{minipage}

\begin{minipage}{\textwidth}
\centering
\includegraphics[width=0.49\textwidth]{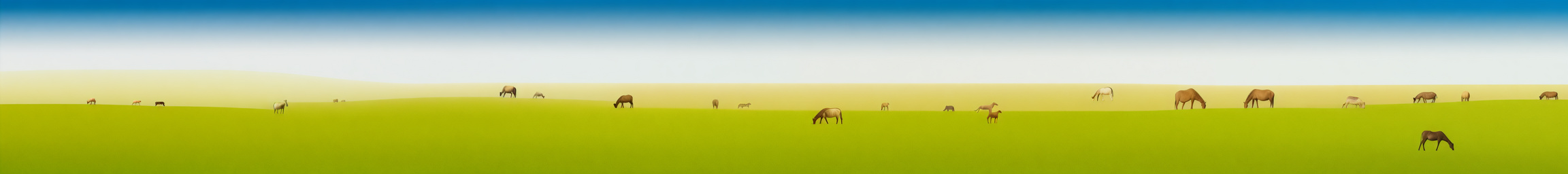}\hfill
\includegraphics[width=0.49\textwidth]{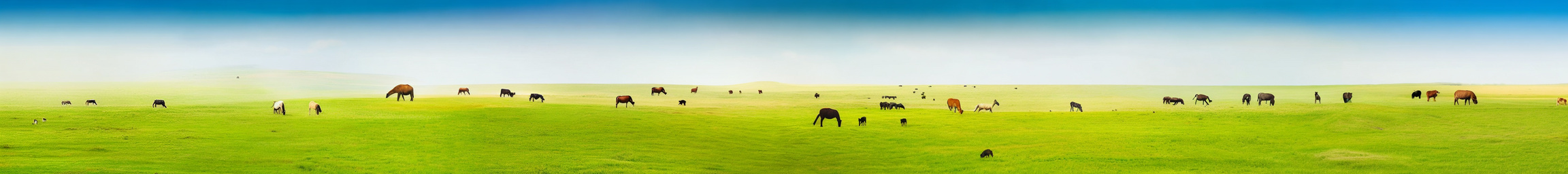}\\[-1pt]
{\tiny \color{gray!80!black}\texttt{grassland}}\\[3pt]
\end{minipage}

\begin{minipage}{\textwidth}
\centering
\includegraphics[width=0.49\textwidth]{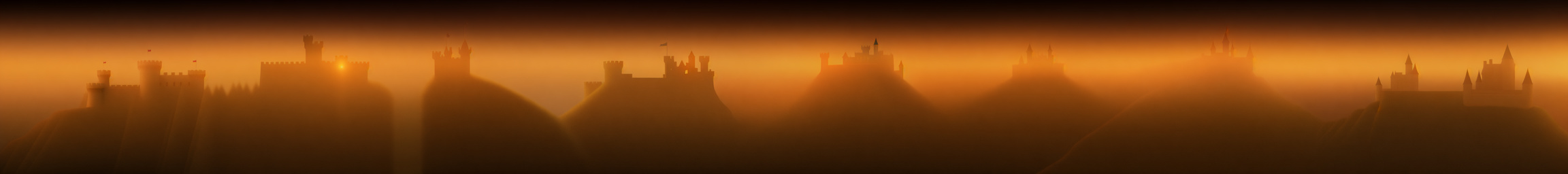}\hfill
\includegraphics[width=0.49\textwidth]{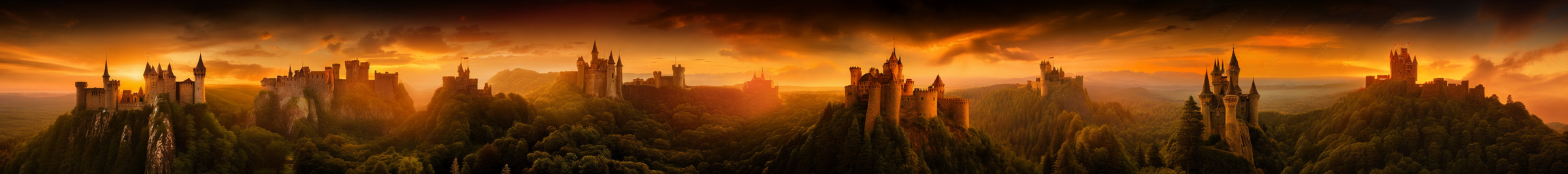}\\[-1pt]
{\tiny \color{gray!80!black}\texttt{castle}}\\[3pt]
\end{minipage}

\begin{minipage}{\textwidth}
\centering
\includegraphics[width=0.49\textwidth]{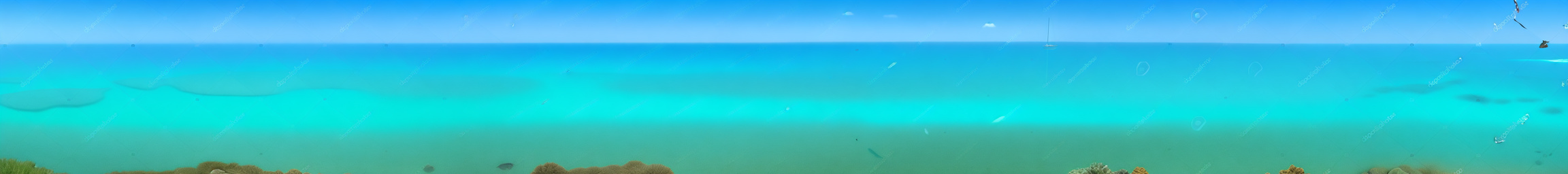}\hfill
\includegraphics[width=0.49\textwidth]{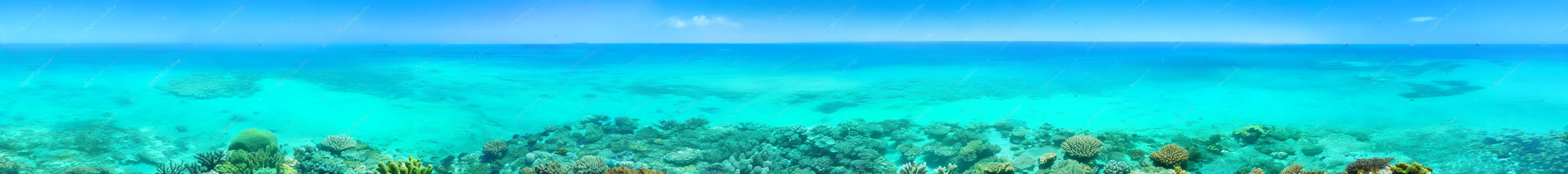}\\[-1pt]
{\tiny \color{gray!80!black}\texttt{coral-reef}}\\[3pt]
\end{minipage}

\begin{minipage}{\textwidth}
\centering
\includegraphics[width=0.49\textwidth]{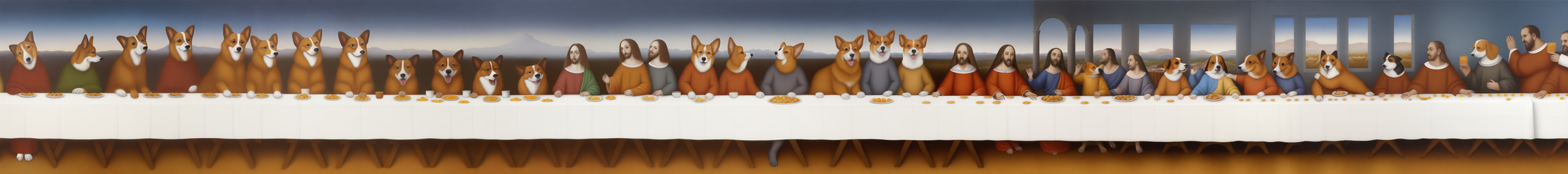}\hfill
\includegraphics[width=0.49\textwidth]{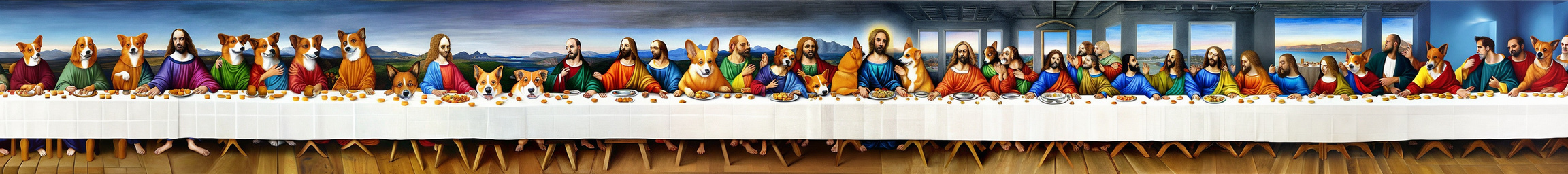}\\[-1pt]
{\tiny \color{gray!80!black}\texttt{corgis}}\\[3pt]
\end{minipage}

\vspace{-2mm}
\caption{\textbf{Panoramic image results.} Per-prompt coarse scaffold and refined output (full prompts in~Appendix~\ref{app:prompt_list}).}
\label{fig:pano_all}
\end{figure}

\paragraph{Metrics.}
\textit{Intra-LPIPS}~\citep{zhang2018unreasonable} measures the mean pairwise LPIPS perceptual distance between all pairs of non-overlapping $512 \times 512$ crops within a single panorama, capturing local style consistency across distant spatial regions. \textit{Intra-Style loss}~\citep{gatys2016image} computes the mean pairwise Gram-matrix distance between VGG features (layers \texttt{relu1\_2}, \texttt{relu2\_2}, \texttt{relu3\_3}, \texttt{relu4\_3}) of these crops, measuring global style coherence. \textit{CLIP score}~\citep{radford2021learning} evaluates text-image alignment between the prompt and the full generated panorama using the ViT-B/32 CLIP model.

\paragraph{Per-prompt results.}
We evaluate on $14$ text prompts with $5$ random seeds each. Figure~\ref{fig:pano_all} shows the coarse scaffold and the refined output for all prompts.

\subsection{Long Video Generation}\label{app:video}

\paragraph{Task description.}
Given a text prompt, the task is to generate a temporally coherent long video ($273$ frames at $720$p) that preserves subject identity and scene structure over a much longer horizon than the pretrained model natively supports. CogVideoX-2B~\citep{yang2025cogvideox} generates $50$-frame clips, so producing a long video requires composing $N = 9$ temporal chunks with $50\%$ overlap.

\paragraph{Guidance.}
As in the image domain, local temporal consistency across adjacent chunks is enforced via the Bethe approximation in~\eqref{eq:bethe approximation} with ramp blending in the shared temporal overlap. At temporal position $\ell$ within an overlap of $L$ frames:
\begin{equation}\label{eq:ramp_blend_video}
\hat{X}^{\texttt{fused}}_{0|t}(\ell) \;=\; w(\ell) \cdot \hat{X}^i_{0|t} \;+\; \bigl(1 - w(\ell)\bigr) \cdot \hat{X}^{i+1}_{0|t},
\qquad
w(\ell) \;=\; 1 - \frac{\ell}{L}.
\end{equation}
Global temporal coherence is achieved through the coarse mean regularization in~\eqref{eq:shared consensus}.

\paragraph{Hyperparameters.}
We use $T = 50$ DDIM steps with classifier-free guidance scale $6.0$. The coarse stage uses $\lambda_0 = 0.2$ with the same linear ramp schedule as the image setting. Unlike the image setting, we use noise-bar directional noise (rather than independently sampled noise) in the coarse stage: all local chunks share a common noise realization in the DDIM direction term, which we find critical for maintaining temporal coherence across chunk boundaries. The coarse mean decay follows a cosine schedule$\lambda_t \;=\; \lambda_0 \cdot \cos\!\left(\frac{\pi\, t}{2T}\right)$,
which provides a smoother transition from strong to weak regularization compared to linear decay. For the refinement stage, we set $t^\star = 0.7T$ with stochasticity $\eta = 1.0$.

\paragraph{Metrics.}
We use VBench~\citep{huang2024vbench} to evaluate long video quality across four dimensions. \textit{Subject consistency} measures the mean pairwise DINO feature similarity of the main subject across uniformly sampled frame pairs, capturing whether the subject's appearance is preserved over the full video duration. \textit{Temporal flickering} evaluates frame-to-frame visual stability by measuring the mean absolute pixel difference between consecutive frames. \textit{Aesthetic quality} applies a learned aesthetic predictor to individual frames and reports the mean score. \textit{Prompt alignment} measures text-video correspondence via CLIP similarity between the text prompt and uniformly sampled frames.

\begin{figure}[h]
\centering
\begin{minipage}{0.49\textwidth}\centering\scriptsize\texttt{Coarse}\end{minipage}\hfill
\begin{minipage}{0.49\textwidth}\centering\scriptsize\texttt{Refinement}\end{minipage}\\[1pt]
{\tiny \color{gray!60!black}%
\makebox[0.49\textwidth]{\hfill \texttt{frame 1} \hfill \texttt{frame 69} \hfill \texttt{frame 137} \hfill \texttt{frame 205} \hfill \texttt{frame 273} \hfill}%
\hfill
\makebox[0.49\textwidth]{\hfill \texttt{frame 1} \hfill \texttt{frame 69} \hfill \texttt{frame 137} \hfill \texttt{frame 205} \hfill \texttt{frame 273} \hfill}}\\[1pt]
\includegraphics[width=0.49\textwidth]{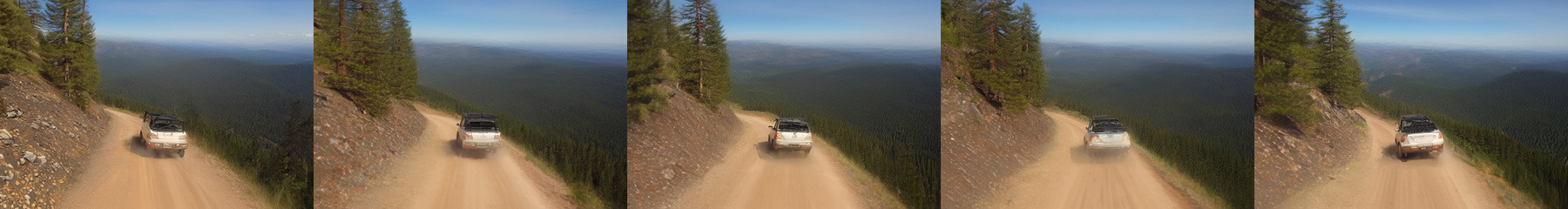}\hfill
\includegraphics[width=0.49\textwidth]{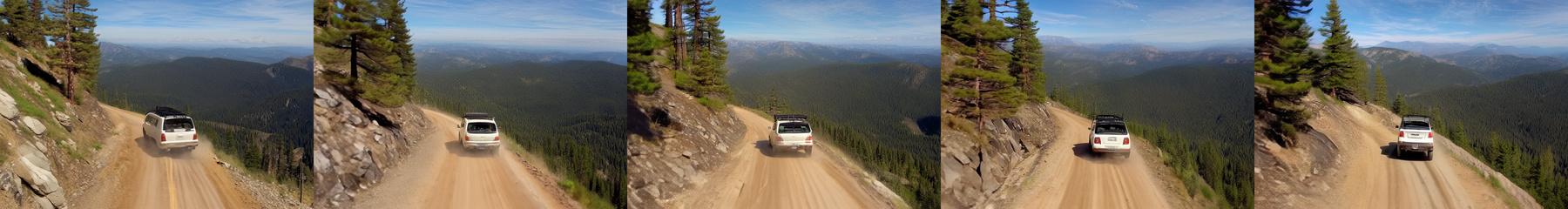}\\[-1pt]
{\tiny \color{gray!80!black}\texttt{suv}}\\[2pt]
\includegraphics[width=0.49\textwidth]{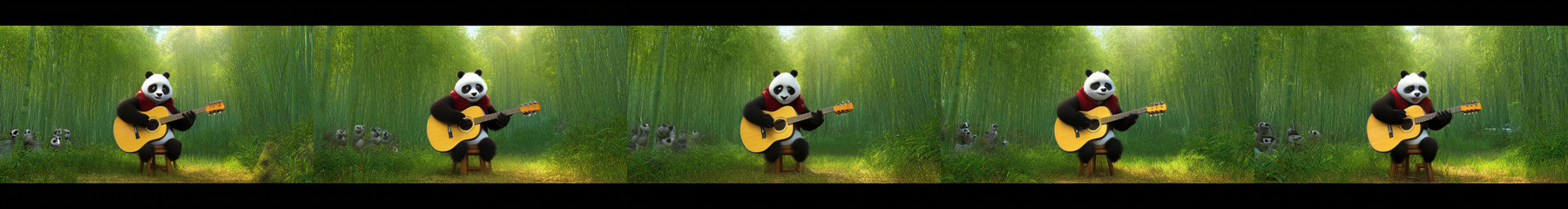}\hfill
\includegraphics[width=0.49\textwidth]{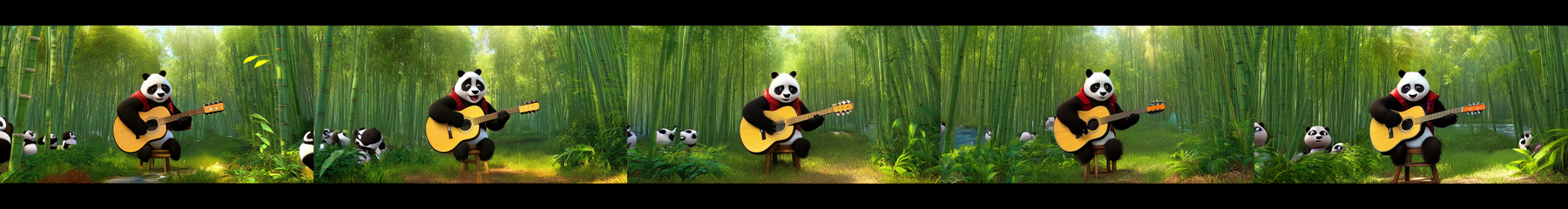}\\[-1pt]
{\tiny \color{gray!80!black}\texttt{panda}}\\[2pt]
\includegraphics[width=0.49\textwidth]{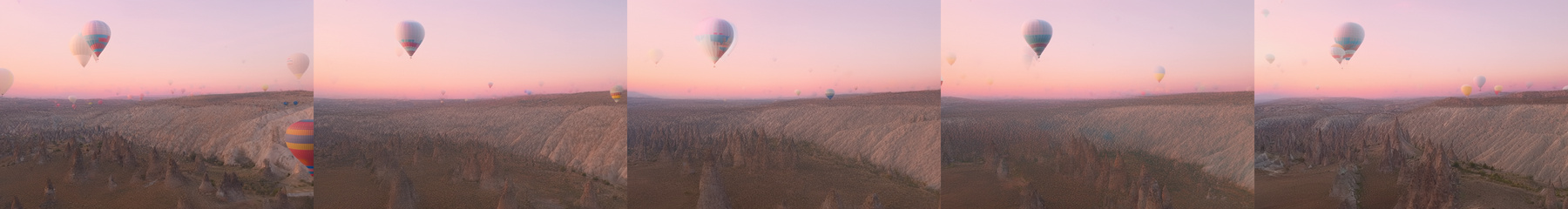}\hfill
\includegraphics[width=0.49\textwidth]{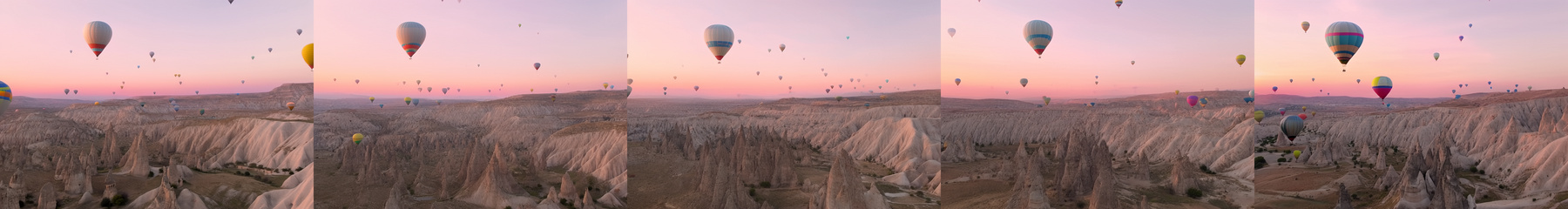}\\[-1pt]
{\tiny \color{gray!80!black}\texttt{balloons}}\\[2pt]
\includegraphics[width=0.49\textwidth]{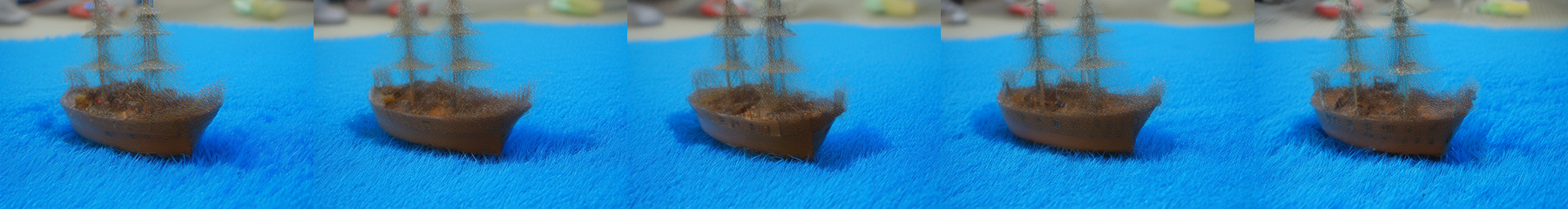}\hfill
\includegraphics[width=0.49\textwidth]{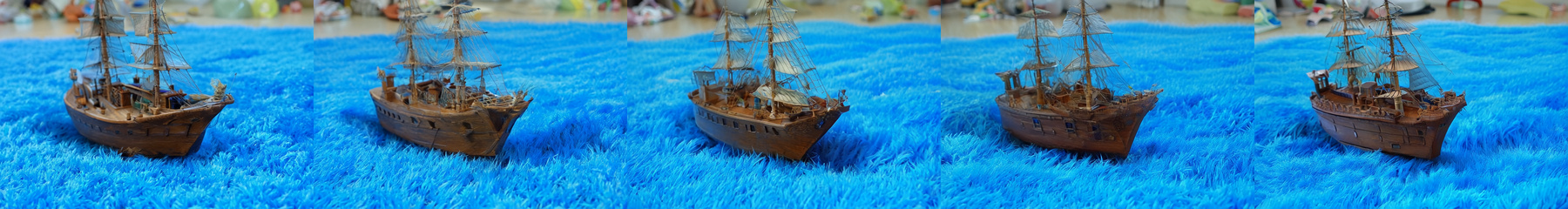}\\[-1pt]
{\tiny \color{gray!80!black}\texttt{toy-ship}}\\[2pt]
\includegraphics[width=0.49\textwidth]{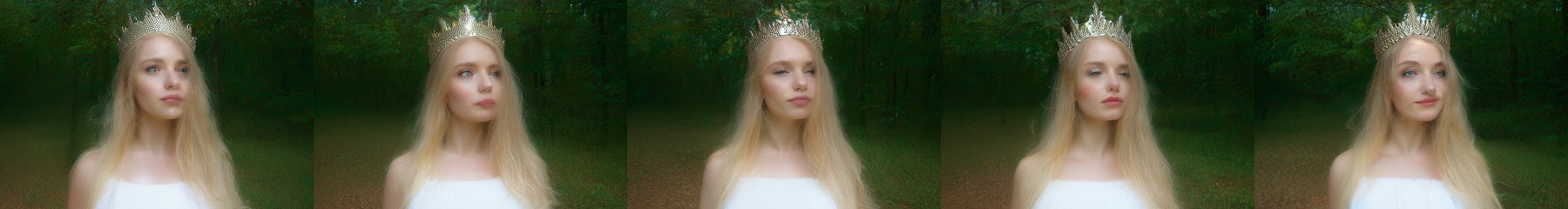}\hfill
\includegraphics[width=0.49\textwidth]{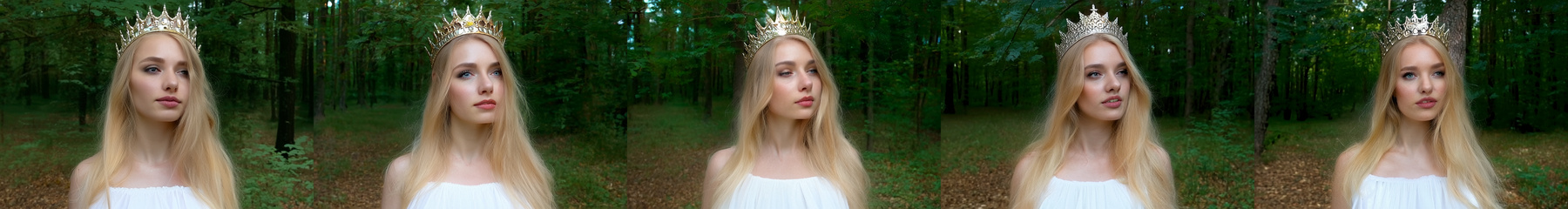}\\[-1pt]
{\tiny \color{gray!80!black}\texttt{young-woman}}\\[2pt]
\includegraphics[width=0.49\textwidth]{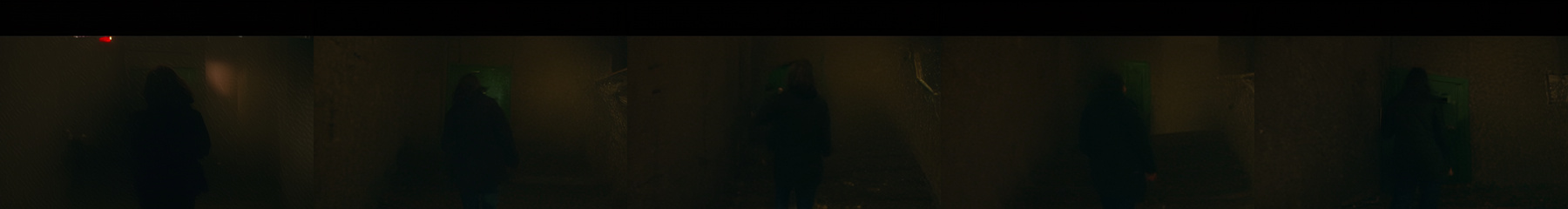}\hfill
\includegraphics[width=0.49\textwidth]{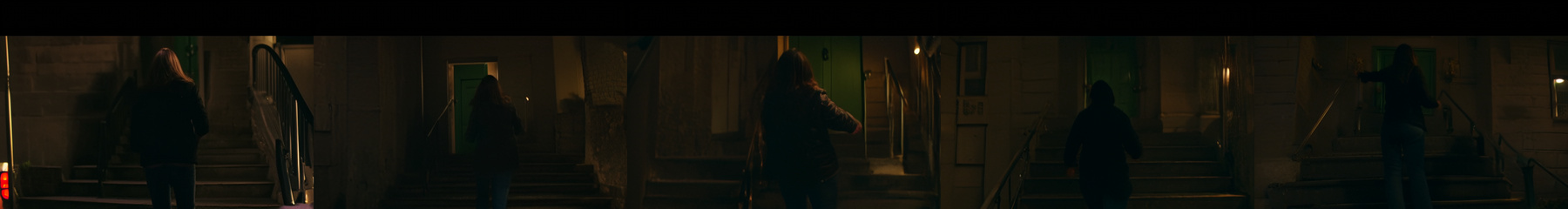}\\[-1pt]
{\tiny \color{gray!80!black}\texttt{white-jeep}}\\[2pt]
\includegraphics[width=0.49\textwidth]{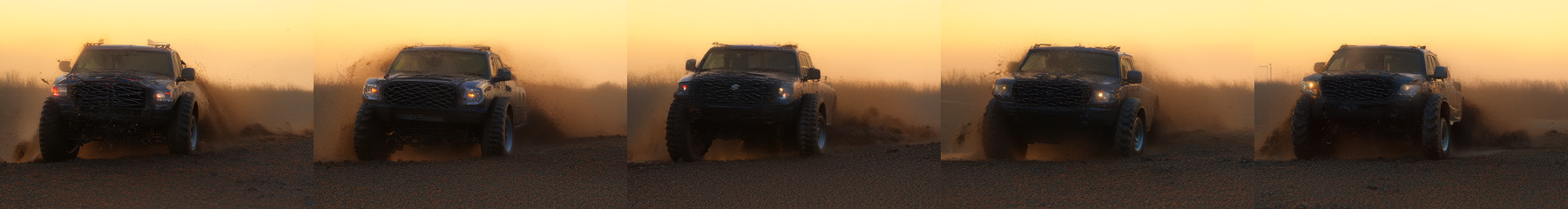}\hfill
\includegraphics[width=0.49\textwidth]{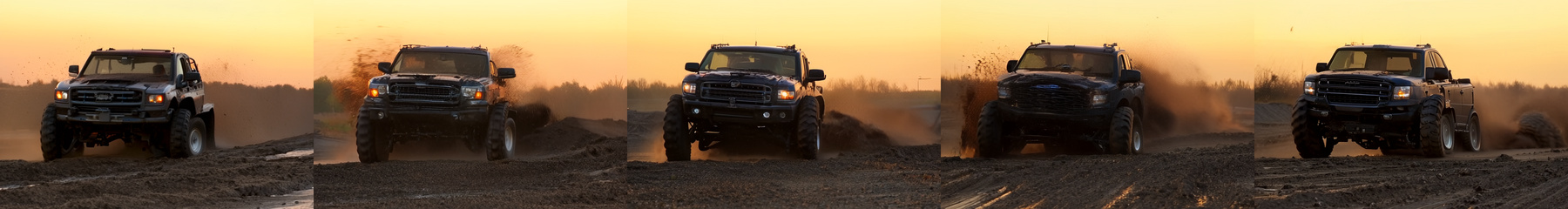}\\[-1pt]
{\tiny \color{gray!80!black}\texttt{F-150}}\\[2pt]
\includegraphics[width=0.49\textwidth]{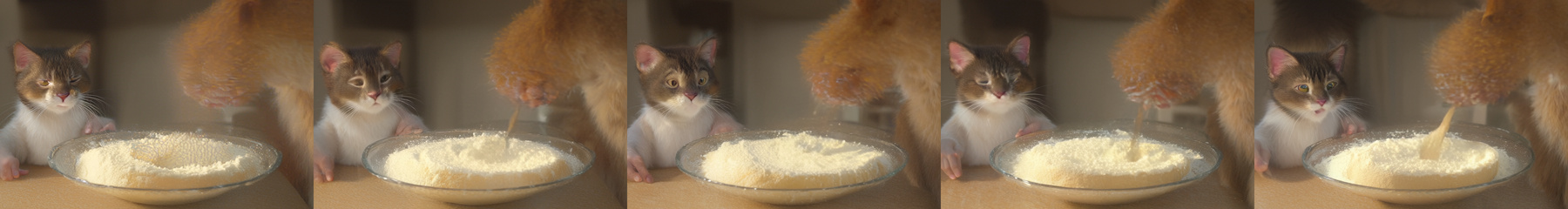}\hfill
\includegraphics[width=0.49\textwidth]{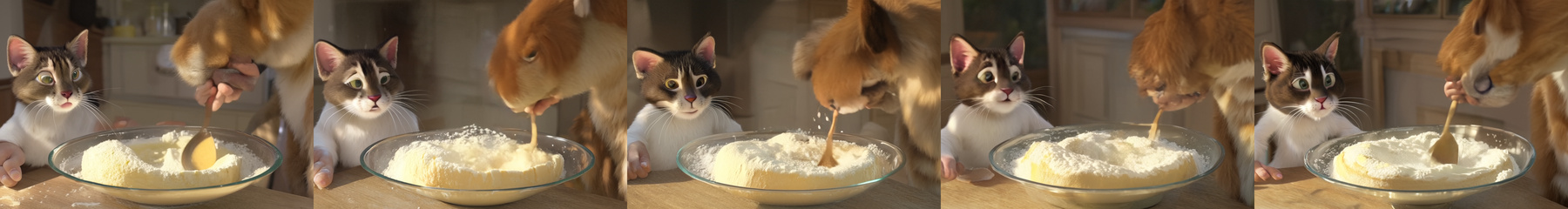}\\[-1pt]
{\tiny \color{gray!80!black}\texttt{baking-cake}}
\vspace{-2mm}
\caption{\textbf{Long video results.} Per-prompt coarse scaffold and refined output. Each strip shows frames 1, 69, 137, 205, 273 of the $273$-frame composed video (full prompts in~Appendix~\ref{app:prompt_list}).}
\label{fig:video_all}
\end{figure}

\paragraph{Per-prompt results.}
We evaluate on $8$ text prompts with $8$ random seeds each. Figure~\ref{fig:video_all} shows sampled frames from the coarse scaffold and the refined output for all prompts.

\subsection{Baseline Configurations}\label{app:baselines}

For all compositional baselines, we use the implementations and hyperparameters from CDGS\footnote{\url{https://github.com/UtkarshMishra04/CDGS_imgvideo} under Apache 2.0 license.}\footnote{\url{https://github.com/UtkarshMishra04/CDGS_ogbench} under MIT license.}~\citep{mishra2026compositional}. GSC~\citep{mishra2023generative} (and its image/video analogues MultiDiffusion~\citep{bar2023multidiffusion} and Gen-L-Video~\citep{wang2023gen}) applies local guidance without resampling or search. SyncDiffusion\footnote{\url{https://github.com/KAIST-Visual-AI-Group/SyncDiffusion} under MIT license.}~\citep{lee2023syncdiffusion} adds a perceptual optimization loop at each denoising step. CDGS~\citep{mishra2026compositional} uses population-based resampling with likelihood-based pruning across inner iterations. 

\subsection{Compute Resources}\label{app:compute}

All experiments are conducted on a single NVIDIA A6000 GPU (48GB VRAM). For robotic planning, composing $N = 9$ segments with $T = 100$ DDIM steps requires approximately $2$~GB of GPU memory. For panoramic image generation, composing a $512 \times 4608$ panorama from $9$ patches with $T = 50$ steps requires approximately $16$~GB. For long video generation, composing $273$ frames from $9$ temporal chunks with $T = 50$ steps requires approximately $23.5$~GB

\subsection{NFE Analysis}\label{app:nfe}

We compare the number of function evaluations (NFE) across methods. Let $T$ denote the total diffusion steps and $N$ the number of local plans.

\begin{itemize}[leftmargin=10pt]
\item \textbf{GSC} : Each denoising step requires one forward pass per local plan. Total NFE $= T \cdot N$.

\item \textbf{CDGS}: Each denoising step involves $K$ inner iterations with a population of size $P$. Total NFE $= K \cdot T \cdot P \cdot N$, where typical values are $K \in [5, 10]$ and $P \in [4, 8]$.

\item \textbf{Ours}: The coarse stage runs $T$ steps and the refinement stage runs $t^\star$ additional steps, each with one forward pass per local plan. Total NFE $= (T + t^\star) \cdot N$. With $t^\star = 0.3T$, this gives $1.3T \cdot N$.
\end{itemize}

The ratio of CDGS to our NFE is $\frac{K \cdot P}{1.3}$. For typical CDGS settings ($K=3$, $P=4$), this gives approximately $9.2\times$, consistent with the $2$--$8\times$ reduction reported in the main text (the exact ratio depends on domain-specific CDGS configurations).

\subsection{Full Prompt Lists}\label{app:prompt_list}

\paragraph{Panoramic image prompts.}
We evaluate on the following $14$ text prompts:
\begin{itemize}[leftmargin=15pt, itemsep=0pt, parsep=1pt]
\item \texttt{landscape}: ``A beautiful landscape with mountains and a river''
\item \texttt{dolomites}: ``A photo of the dolomites with red lava flowing through the valley''
\item \texttt{babbling-brook}: ``A photo of lush forest with a babbling brook''
\item \texttt{misty-fog}: ``A photo of a forest with a misty fog''
\item \texttt{cartoon}: ``Cartoon panorama of spring summer beautiful nature''
\item \texttt{skiers}: ``A photo of a snowy mountain peak with skiers''
\item \texttt{anime}: ``Natural landscape in anime style illustration''
\item \texttt{futuristic}: ``Skyline of a futuristic city with flying cars''
\item \texttt{silhouette}: ``Silhouette wallpaper of a dreamy scene with shooting stars''
\item \texttt{palm-trees}: ``A beach with palm trees''
\item \texttt{grassland}: ``A photo of a grassland with animals''
\item \texttt{castle}: ``A cinematic view of a castle in the sunset''
\item \texttt{coral-reef}: ``A photo of a beautiful ocean with coral reef''
\item \texttt{corgis}: ``last supper with cute corgis''
\end{itemize}

\paragraph{Long video prompts.}
We evaluate on the following $8$ text prompts:
\begin{itemize}[leftmargin=15pt, itemsep=1pt, parsep=1pt]
\item \texttt{suv}: ``The camera follows behind a white vintage SUV with a black roof rack as it speeds up a steep dirt road surrounded by pine trees on a steep mountain slope, dust kicks up from its tires, the sunlight shines on the SUV as it speeds along the dirt road, casting a warm glow over the scene. The dirt road curves gently into the distance, with no other cars or vehicles in sight. The trees on either side of the road are redwoods, with patches of greenery scattered throughout. The car is seen from the rear following the curve with ease, making it seem as if it is on a rugged drive through the rugged terrain. The dirt road itself is surrounded by steep hills and mountains, with a clear blue sky above with wispy clouds.''
\item \texttt{panda}: ``A cute happy panda, dressed in a small, red jacket and a tiny hat, sits on a wooden stool in a serene bamboo forest. The panda's fluffy paws strum a miniature acoustic guitar, producing soft, melodic tunes, move hands, singings. Nearby, a few other pandas gather, watching curiously and some clapping in rhythm. Sunlight filters through the tall bamboo, casting a gentle glow on the scene. The panda's face is expressive, showing concentration and joy as it plays. The background includes a small, flowing stream and vibrant green foliage, enhancing the peaceful and magical atmosphere of this unique musical performance.''
\item \texttt{balloons}: ``A group of colorful hot air balloons take off at dawn in Cappadocia, Turkey. Dozens of balloons in various bright colors and patterns slowly rise into the pink and orange sky. Below them, the unique landscape of Cappadocia unfolds, with its distinctive `fairy chimneys' -- tall, cone-shaped rock formations scattered across the valley. The rising sun casts long shadows across the terrain, highlighting the otherworldly topography.''
\item \texttt{toy-ship}: ``A detailed wooden toy ship with intricately carved masts and sails is seen gliding smoothly over a plush, blue carpet that mimics the waves of the sea. The ship's hull is painted a rich brown, with tiny windows. The carpet, soft and textured, provides a perfect backdrop, resembling an oceanic expanse. Surrounding the ship are various other toys and children's items, hinting at a playful environment. The scene captures the innocence and imagination of childhood, with the toy ship's journey symbolizing endless adventures in a whimsical, indoor setting.''
\item \texttt{young-woman}: ``A young woman with beautiful and clear eyes and blonde hair standing and white dress in a forest wearing a crown. She seems to be lost in thought, and the camera focuses on her face. The video is of high quality, and the view is very clear. High quality, masterpiece, best quality, highres, ultra-detailed, fantastic.''
\item \texttt{white-jeep}: ``A woman walks away from a white Jeep parked on a city street at night, then ascends a staircase and knocks on a door. The woman, wearing a dark jacket and jeans, walks away from the Jeep parked on the left side of the street, her back to the camera; she walks at a steady pace, her arms swinging slightly by her sides; the street is dimly lit, with streetlights casting pools of light on the wet pavement; the camera follows the woman from behind as she walks up a set of stairs towards a building with a green door; she reaches the top of the stairs and turns left, continuing to walk towards the building; she reaches the door and knocks on it with her right hand.''
\item \texttt{F-150}: ``At sunset, a modified Ford F-150 Raptor roared past on the off-road track. The raised suspension allowed the huge explosion-proof tires to flip freely on the mud, and the mud splashed on the roll cage.''
\item \texttt{baking-cake}: ``A cat and a dog baking a cake together in a kitchen. The cat is carefully measuring flour, while the dog is stirring the batter with a wooden spoon. The kitchen is cozy, with sunlight streaming through the window.''
\end{itemize}

\newpage

\end{document}